\documentclass[10pt,twocolumn,letterpaper]{article}

\usepackage{cvpr}

\makeatletter
\@namedef{ver@everyshi.sty}{}
\makeatother

\usepackage{graphicx}
\usepackage{amsmath}
\usepackage{amssymb}
\usepackage{booktabs}
\usepackage{comment}
\usepackage{color}
\usepackage{tabularx}
\usepackage{colortbl}
\usepackage{tabu}
\usepackage{multirow}
\usepackage{setspace}
\usepackage[font=small,labelsep=period]{caption}
\usepackage[dvipsnames]{xcolor}
\usepackage{tikz}
\usepackage{pgfplots}
\usepackage{pgfkeys}
\usepackage{xcolor}
\usepackage{xspace}
\usepackage{cuted}

\usepackage[moderate]{savetrees}

\definecolor{bblue}{rgb}{0.0,0.2,0.8}
\definecolor{myred}{rgb}{0.9,0.1,0.1}
\definecolor{ccol}{rgb}{0.91,0.91,0.91}
\usepackage[pagebackref=true,breaklinks=true,colorlinks=true,urlcolor=magenta,bookmarks=false,citecolor=bblue]{hyperref}

\usepackage[accsupp]{axessibility} %

\newcolumntype{Y}{>{\centering\arraybackslash}X}
\newcolumntype{R}{>{\raggedleft\arraybackslash}X}
\newcolumntype{L}{>{\raggedright\arraybackslash}X}

\newcommand\customparagraph[1]{\vspace{0.7em}\noindent\textbf{#1}}
\newcommand{\mytilde}{\raise.17ex\hbox{$\scriptstyle\sim$}}

\def\addlegendimage{\csname pgfplots@addlegendimage\endcsname}

\usepackage[capitalize]{cleveref}
\crefname{section}{Sec.}{Secs.}
\Crefname{section}{Section}{Sections}
\Crefname{table}{Table}{Tables}
\crefname{table}{Tab.}{Tabs.}

\makeatletter
\renewcommand{\fnum@figure}{Figure \thefigure}
\makeatother

\def\cvprPaperID{X}
\def\confName{CVPR}
\def\confYear{2025}

\usepackage{cuted}
\usepackage{capt-of}

\begin{document}

\title{Introducing HOT3D: An Egocentric Dataset for 3D Hand and Object Tracking}

\newcommand{\namesep}{, }

\author{
Prithviraj Banerjee\namesep
Sindi Shkodrani\namesep
Pierre Moulon\namesep
Shreyas Hampali\namesep
Fan Zhang\namesep
Jade Fountain,\vspace{0.2em}\\
Edward Miller\namesep
Selen Basol\namesep
Richard Newcombe\namesep
Robert Wang\namesep
Jakob Julian Engel\namesep
Tomas Hodan\vspace{9.5pt} \\
{\normalsize Meta Reality Labs\hspace{1.8em}\href{https://facebookresearch.github.io/hot3d/}{facebookresearch.github.io/hot3d}}
}

\maketitle

\begin{strip}
\begin{minipage}{\textwidth}\centering
\vspace{-30.5pt}
\centering
\begin{minipage}{0.195\linewidth}%
\includegraphics[width=\linewidth]{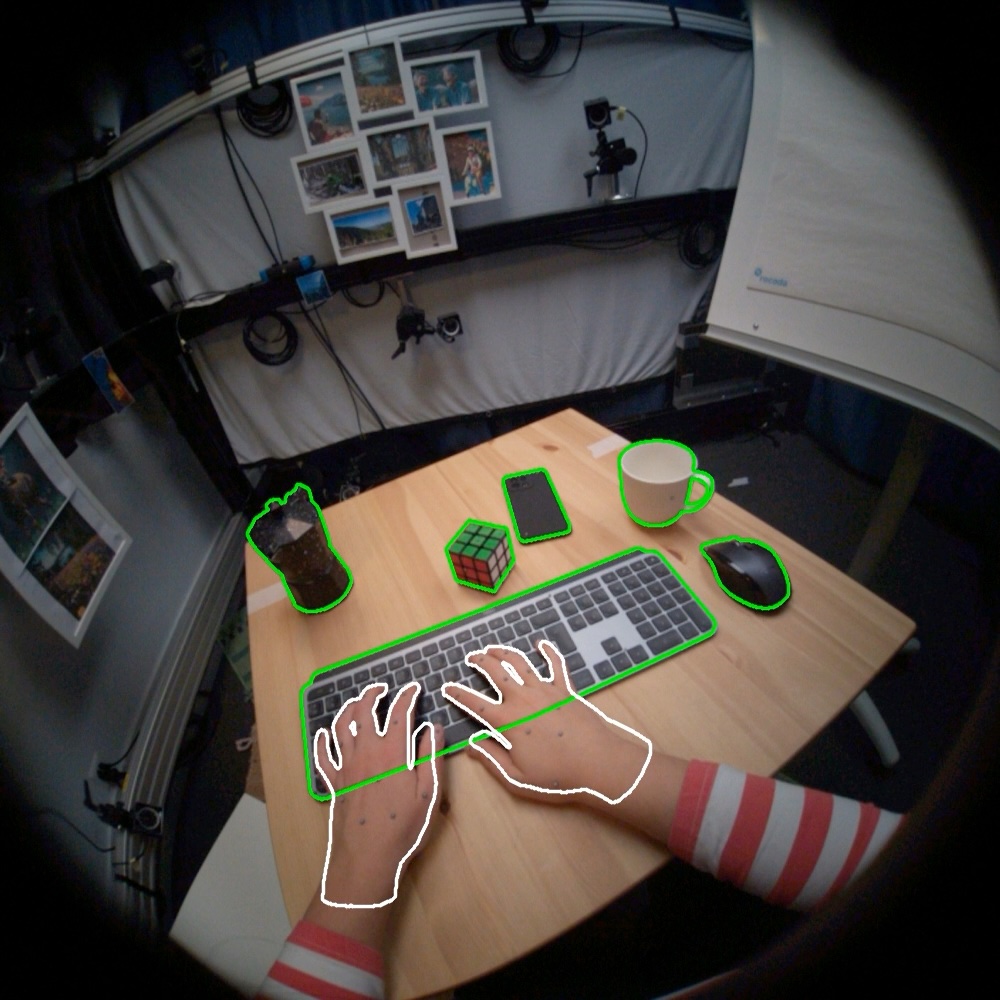}\\[0.2mm]
\includegraphics[width=0.49\linewidth]{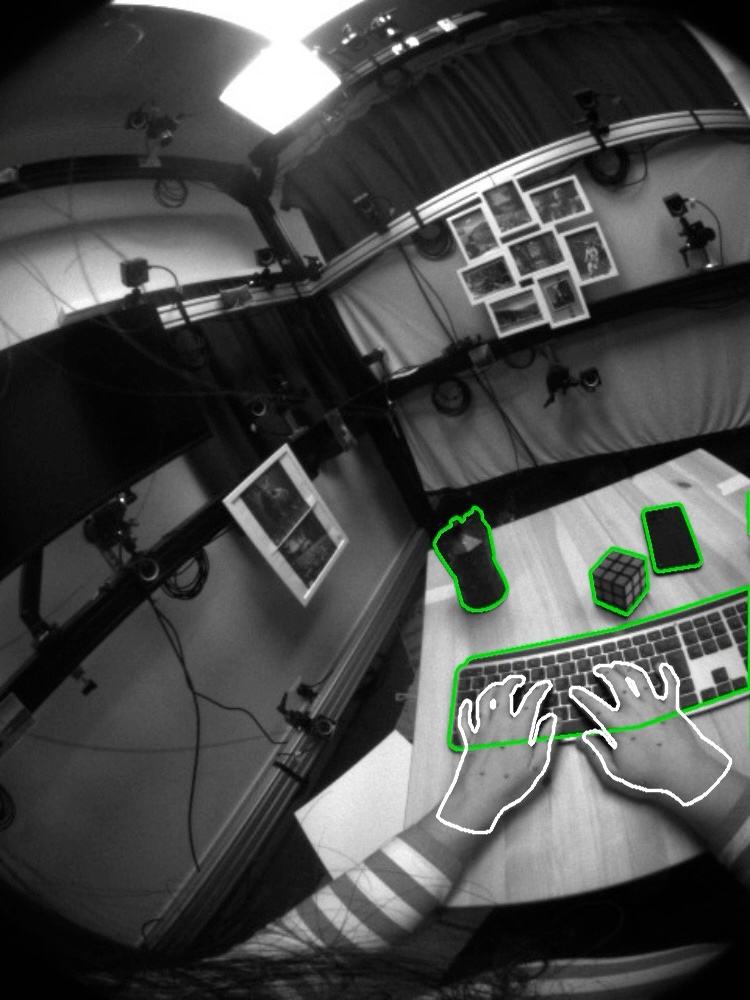}
\includegraphics[width=0.49\linewidth]{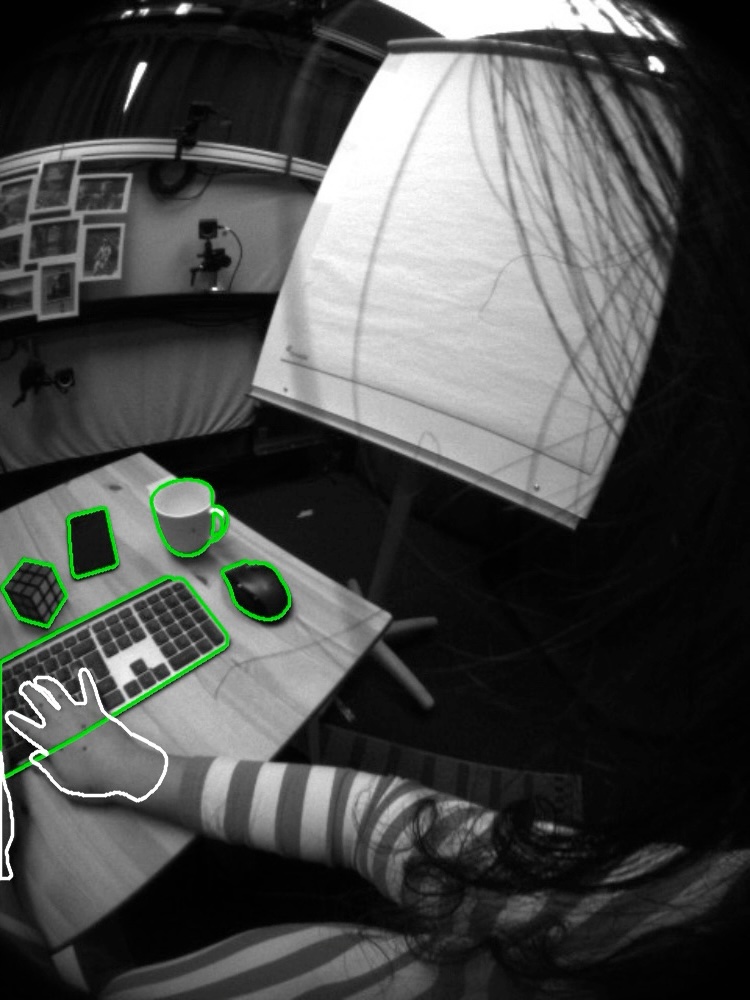}%
\end{minipage}\hfill%
\begin{minipage}{0.195\linewidth}%
\includegraphics[width=\linewidth]{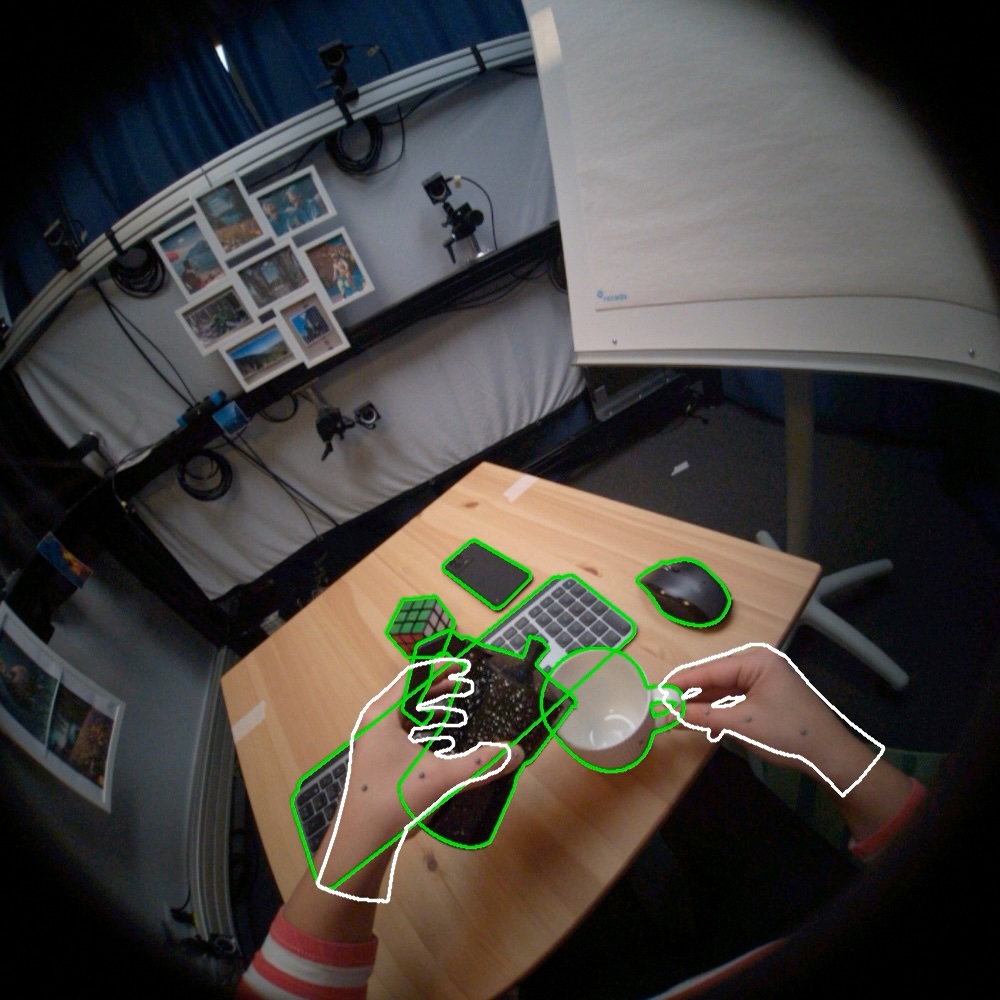}\\[0.2mm]
\includegraphics[width=0.49\linewidth]{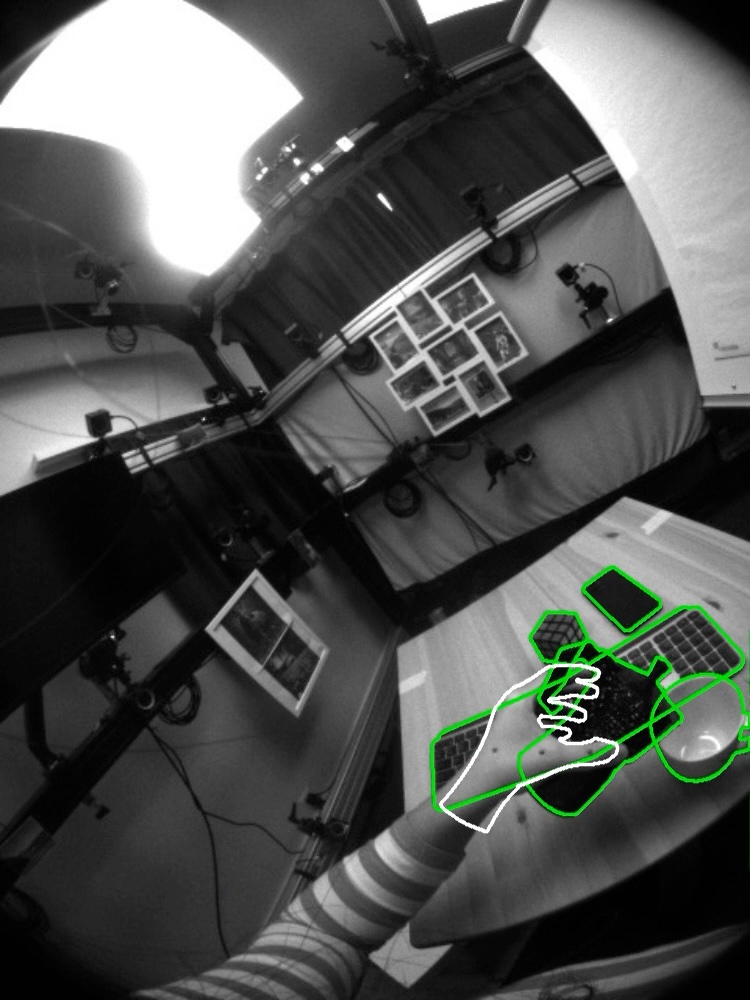}
\includegraphics[width=0.49\linewidth]{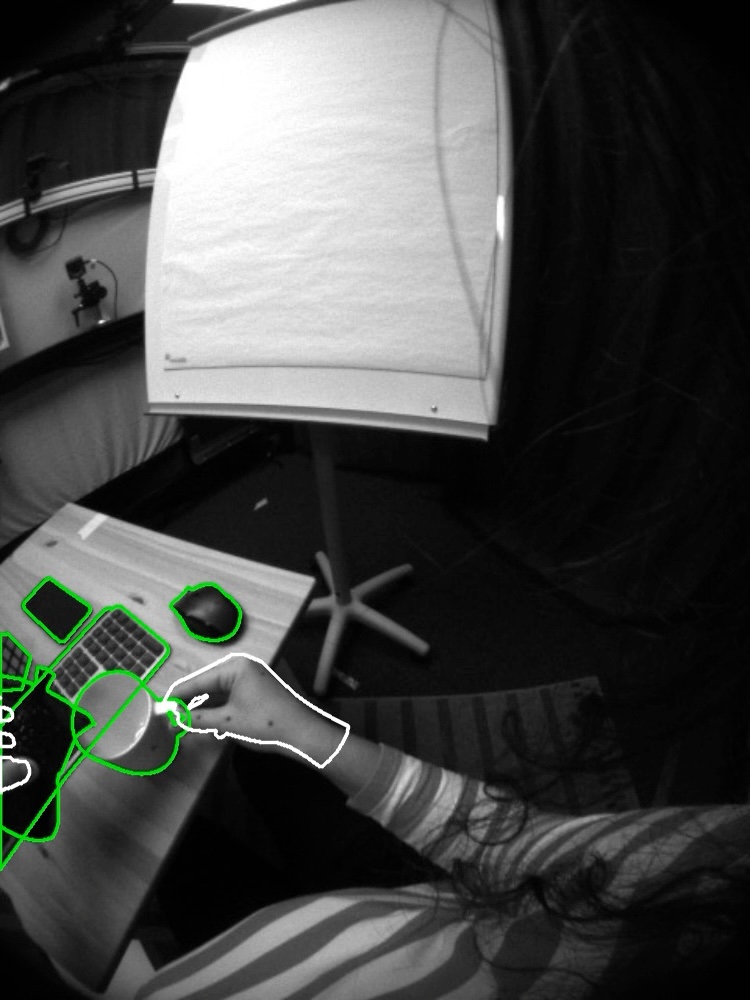}%
\end{minipage}\hfill%
\begin{minipage}{0.195\linewidth}%
\includegraphics[width=\linewidth]{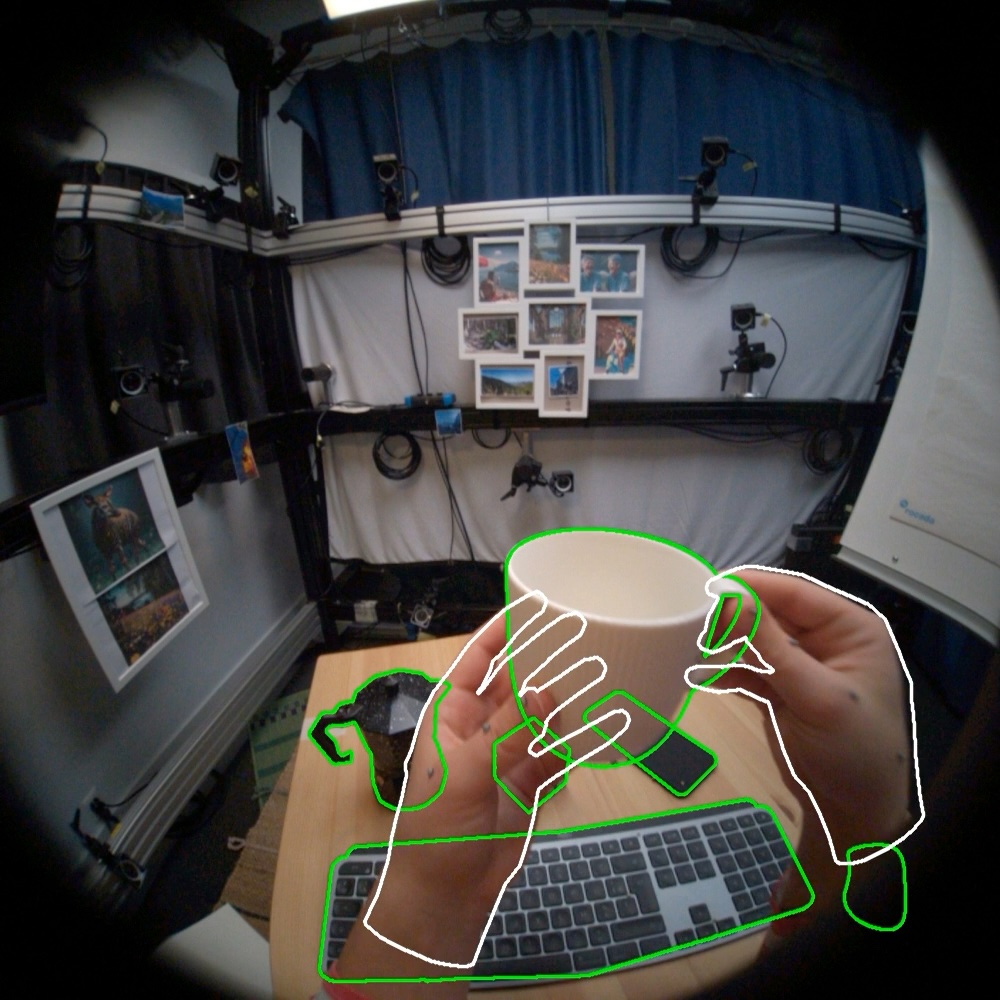}\\[0.2mm]
\includegraphics[width=0.49\linewidth]{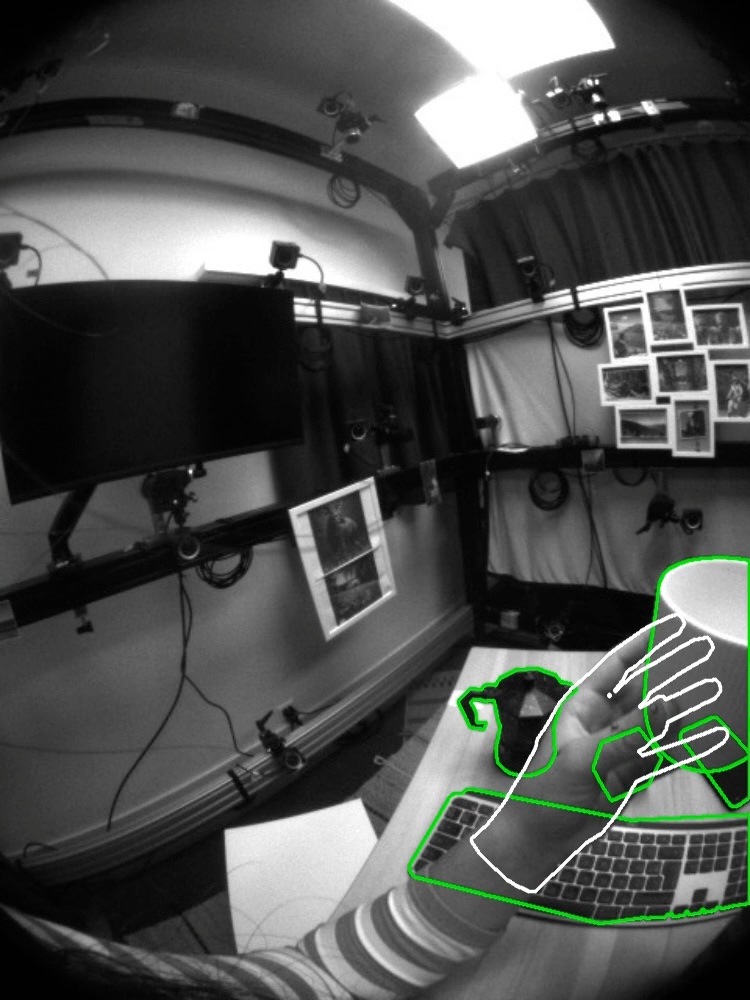}
\includegraphics[width=0.49\linewidth]{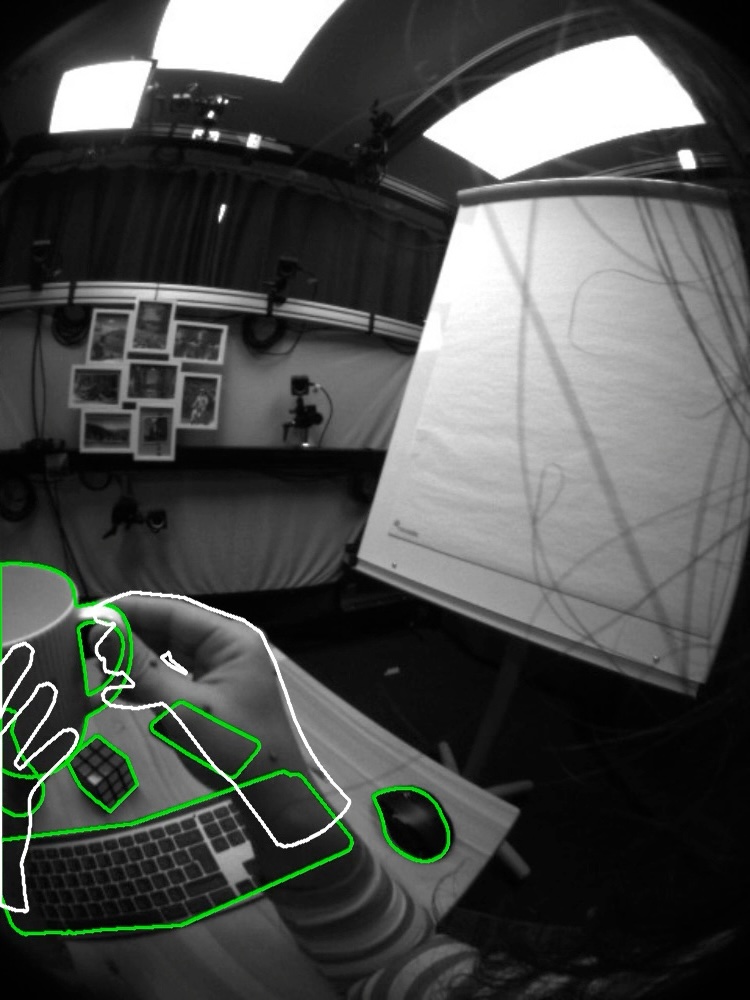}%
\end{minipage}\hfill%
\begin{minipage}{0.393\linewidth}%
{\setlength{\fboxsep}{0pt}\setlength{\fboxrule}{0.5pt}
\fbox{\includegraphics[width=\linewidth]{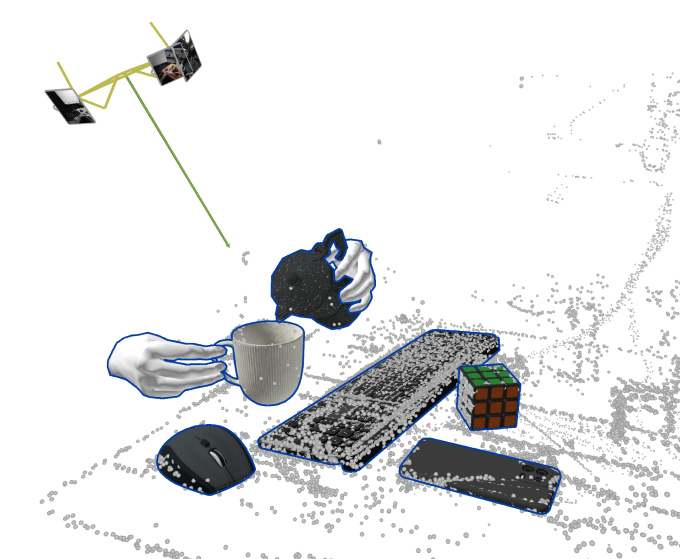}}}%
\end{minipage}
\captionof{figure}{
    \textbf{HOT3D overview.} The dataset includes multi-view egocentric 
    image streams from Aria~\cite{engel2023project} and Quest 3~\cite{Quest3}
    annotated with high-quality ground-truth 3D poses and models of hands and objects.
    Three multi-view frames from Aria are shown on the left, with contours of 3D models of hands and objects in the ground-truth poses in white and green, respectively.
    Aria also provides 3D point clouds from SLAM (right) and eye gaze information.
    Hand annotations are provided in the UmeTrack~\cite{han2022umetrack} and MANO~\cite{mano:romero:tog17} formats, while objects are represented by 3D mesh models from an in-house scanner.
}
\label{fig:overview}
\end{minipage}
\vspace{5pt}
\end{strip}

\makebox[0pt][c]{%
\hspace{0.925\columnwidth}
\begin{minipage}[b]{\columnwidth}
\vspace{2pt}
\begin{abstract}
\vspace{-4pt}
We introduce HOT3D, a publicly available dataset for egocentric hand and object tracking in 3D.
The dataset offers over 833 minutes (more than 3.7M images) of multi-view RGB/monochrome image streams showing 19 subjects interacting with 33 diverse rigid objects, multi-modal signals such as eye gaze or scene point clouds, as well as comprehensive ground truth annotations including 3D poses of objects, hands, and cameras, and 3D models of hands and objects.
In addition to simple pick-up/observe/put-down actions, HOT3D contains scenarios resembling typical actions in a kitchen, office, and living room environment. The dataset is recorded by two head-mounted devices from Meta: Project Aria, a research prototype of light-weight AR/AI glasses, and Quest~3, a production VR headset sold in millions of units. Ground-truth poses were obtained by a professional motion-capture system using small optical markers attached to hands and objects.
Hand annotations are provided in the UmeTrack and MANO formats and objects are represented by 3D meshes with PBR materials obtained by an in-house scanner.
We aim to accelerate research on egocentric hand-object interaction by making the HOT3D dataset publicly available and by co-organizing public challenges on the dataset at ECCV 2024.
The dataset can be downloaded from the project website:
\href{https://facebookresearch.github.io/hot3d/}{facebookresearch.github.io/hot3d}.
\end{abstract}%
\end{minipage}
}%

\section{Introduction}

We use our hands to communicate with others, interact with objects, or utilize objects as tools to act upon other objects. The dexterity with which we can manipulate objects is unmatched by other species and has been a key factor in our evolution~\cite{bardo2022precision}.
Hand-object interaction has therefore naturally received a considerable attention of various research fields, including computer vision~\cite{oikonomidis2018hands18}.

A vision-based system for automatic understanding of~hand-object interaction, which would be able to capture information about 3D motion, shape and contact of hands and objects, will unlock new types of applications. For example, such a system will enable transferring manual skills between users by first capturing expert users performing a sequence of hand-object interactions (when assembling a piece of furniture, doing a tennis serve, \etc), and by using the captured information to guide less experienced users, \eg, via AR glasses. The skills could be similarly transferred from humans to robots, enabling autonomous robots that can learn on the fly. The system could also help an AI assistant to better understand the context of a user's actions or enable new input capabilities for AR/VR users, \eg, by turning any physical surface to a virtual keyboard, or any pencil to a multi-functional magic wand. However, the accuracy and speed of existing methods for understanding hand-object~interaction are not sufficient to reliably support such applications.

To accelerate computer vision research on hand-object~interaction, we are publicly releasing HOT3D, an egocentric dataset for training and evaluating methods for hand and object tracking in 3D. The dataset includes over 833 minutes of egocentric image sequences, which include over 1.5M multi-view frames (over 3.7M images) and show 19 subjects interacting with 33 diverse rigid objects. Besides a simple inspection scenario, where subjects pick up, observe, and put down the objects, the sequences show scenarios resembling typical actions in the kitchen, office, and living room.

\begin{figure}[t]
  \centering
  \includegraphics[width=1.0\linewidth]{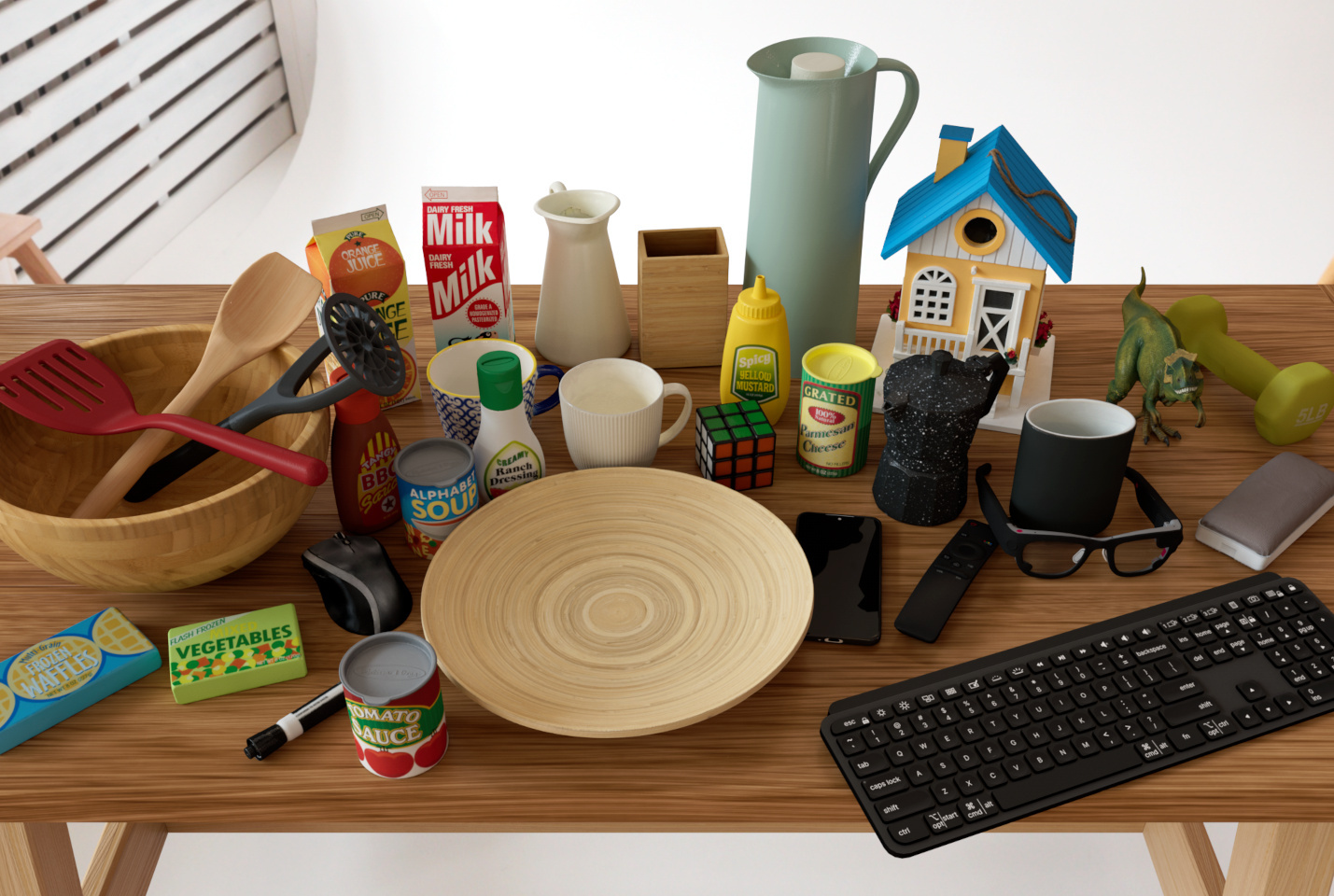}
  
   \caption{\textbf{High-quality 3D mesh models.} This image shows a rendering of the 33 object models, demonstrating their quality. The models were obtained by an in-house 3D scanner and include PBR materials, which enable rendering of photo-realistic training images for methods that require it. The collection includes household and office objects of diverse appearance, size, and affordances.
   }
   \label{fig:models}
\end{figure}

HOT3D is recorded by two recent head-mounted devices from Meta: Project Aria~\cite{engel2023project}, which is a research prototype of light-weight AI glasses, and Quest~3~\cite{Quest3}, which is a production virtual-reality headset that has been sold in millions of units.
Hands and objects are annotated with accurate 3D poses collected using a marker-based motion-capture system.
The dataset also offers 3D object models which were obtained by an in-house scanner and include high-resolution geometry and PBR materials~\cite{mcdermott2018pbr} (Fig.~\ref{fig:models}). Recordings from Aria additionally include 3D scene point clouds from SLAM and eye gaze information.

The dataset is primarily intended for training and evaluating methods for model-based and model-free tracking of hands and objects \textbf{in 3D}, and on \textbf{localized, multi-camera video streams} as supposed to monocular views or individual images. For the model-free object tracking setup, the dataset offers on-boarding sequences showing different views at each object. Since images from all streams are synchronized (\ie, captured at the same timestamp), the dataset also enables developing methods that can leverage multi-view and/or temporal information. Furthermore, the dataset can be used for tasks such as 3D object reconstruction and 2D detection or segmentation of hand/object interactions. We also encourage research that leverages the eye gaze information from Project Aria, which can allow to predict the user's intent, or to efficiently allocate computational budget via foveated sensing.

Compared to existing datasets reviewed in Sec.~\ref{sec:related_work}, HOT3D~is unique in its offering of (1) 3.7M multi-view egocentric images recorded with RGB/monochrome cameras from actual headsets, (2) high-quality ground-truth poses of multiple objects, the headset, as well as pose and shape of both hands, (3) non-trivial hand-object interaction scenarios with dynamic grasps, and (4) 3D object models with PBR materials, enabling synthesis of photo-realistic training images. We provide details about the HOT3D dataset and our data collection procedure in Sec.~\ref{sec:dataset}, and call for participation in public challenges on HOT3D, which we co-organize at ECCV 2024, in Sec.~\ref{sec:challenges}.

\begin{figure*}[t]
    \centering
    \setlength{\tabcolsep}{1pt} %
    \renewcommand{\arraystretch}{0.6} %

    \begin{tabular}{cccccc}
        \includegraphics[width=0.164\textwidth]{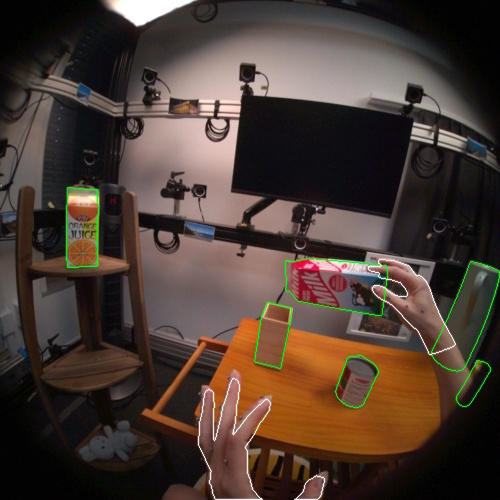} &
        \includegraphics[width=0.164\textwidth]{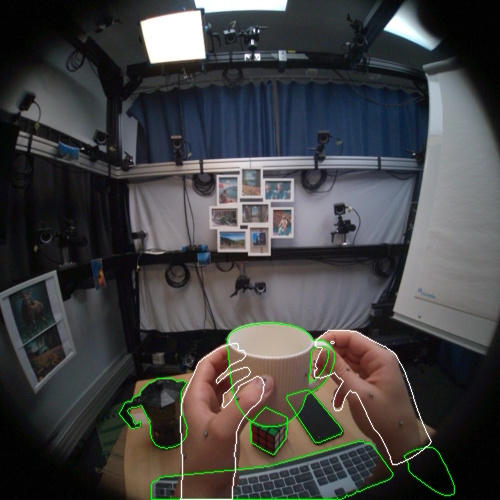} &
        \includegraphics[width=0.164\textwidth]{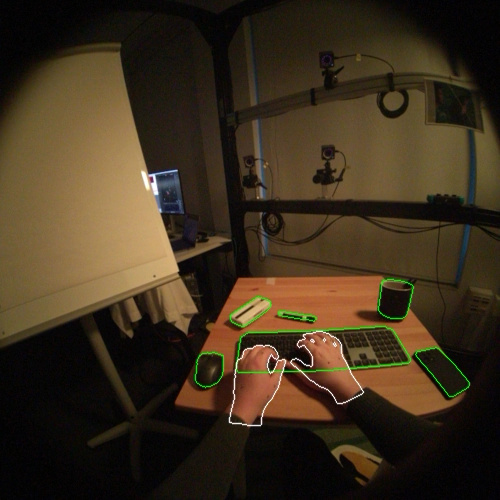} &
        \includegraphics[width=0.164\textwidth]{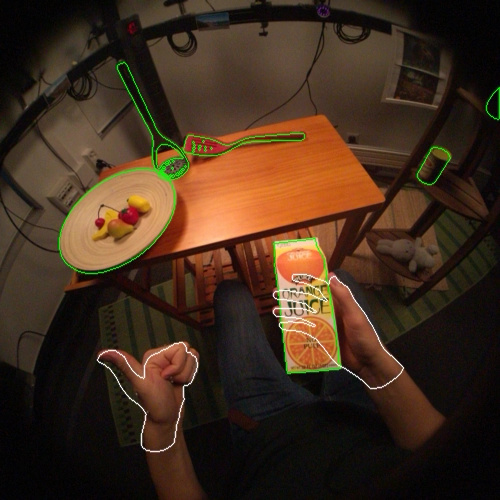} &
        \includegraphics[width=0.164\textwidth]{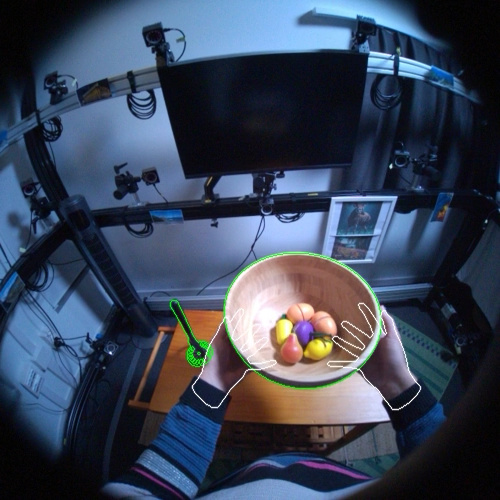} &
        \includegraphics[width=0.164\textwidth]{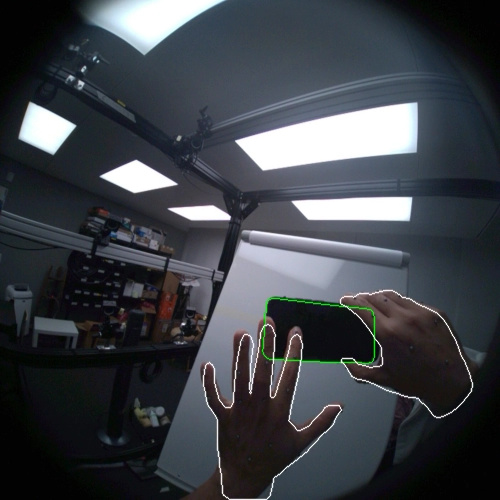} \\
        \includegraphics[width=0.164\textwidth]{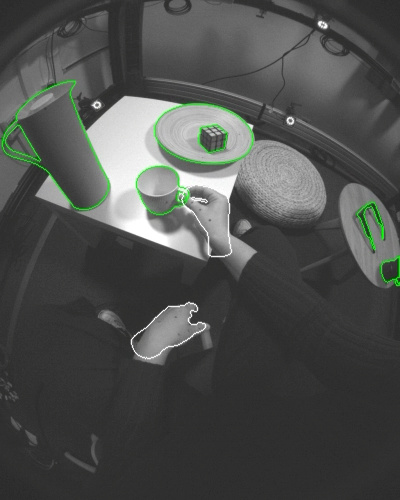} &
        \includegraphics[width=0.164\textwidth]{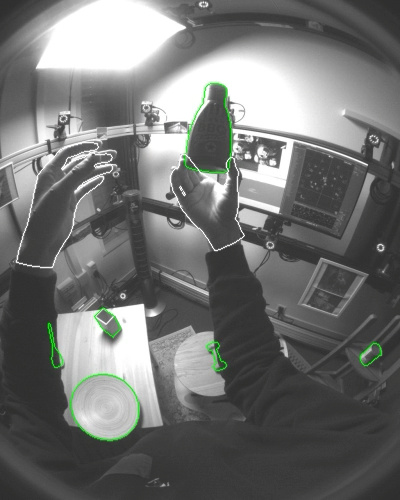} &
        \includegraphics[width=0.164\textwidth]{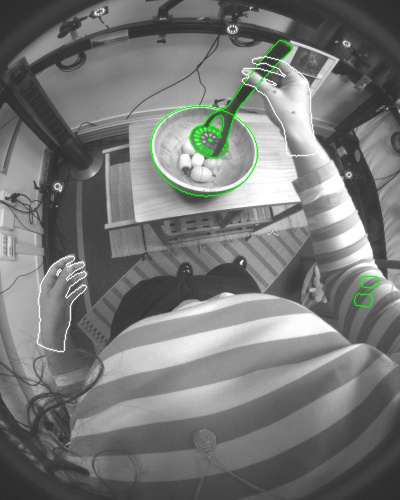} &
        \includegraphics[width=0.164\textwidth]{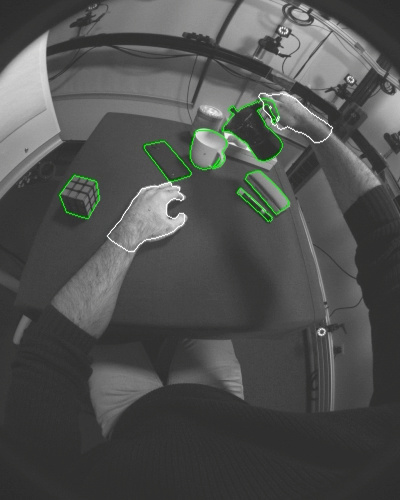} &
        \includegraphics[width=0.164\textwidth]{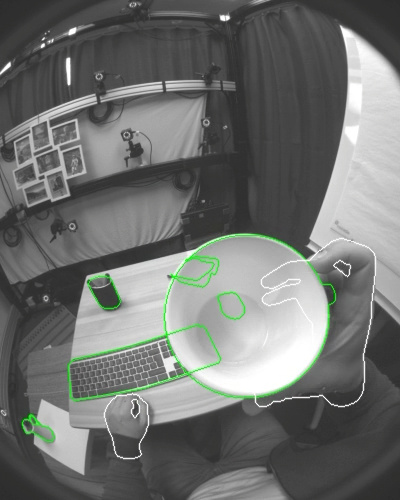} &
        \includegraphics[width=0.164\textwidth]{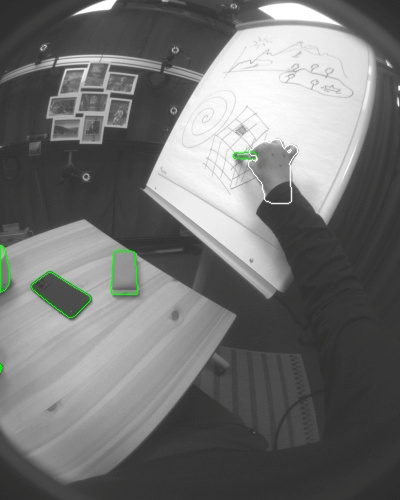}
    \end{tabular}
    \vspace{-0.7ex}

    \caption{\textbf{Sample images from Aria (top) and Quest 3 (bottom).} Aria recordings include one RGB and two monochrome image streams, while Quest 3 recordings include two monochrome streams -- only images from one of the multi-view streams are shown. Contours of 3D models of hands and objects in the ground-truth poses are shown in white and green respectively. In addition to simple pick-up/observe/put-down actions, the subjects perform actions that are common in a kitchen, office, and living room. To increase diversity, the lighting, furniture, and decorations in the capture lab were regularly randomized.}
    \label{fig:sample_data}
    \vspace{1.0ex}
\end{figure*}

\section{Related work} \label{sec:related_work}

The progress of research in computer vision has been strongly influenced by benchmark datasets~\cite{scharstein2002taxonomy,everingham2010pascal,russakovsky2015imagenet,geiger2012we,hodan2018bop} which enable to compare methods and better understand their limitations. In this section, we first review existing datasets with either hand or object pose annotations, and then focus on datasets that offer annotations of hands and hand-manipulated objects.

\customparagraph{Datasets with hands only.}
Vision-based 3D hand pose estimation and tracking has been extensively studied for many years, with the first methods focusing on custom datasets with monochrome images~\cite{rehg1994visual,heap1996towards}. Significant improvements in pose accuracy were later achieved on RGB-D images from datasets such as NYU~\cite{tompson2014real}, ICVL~\cite{tang2014latent}, MSRA~\cite{sun2015cascaded}, Tzionas~\etal \cite{tzionas2016capturing}, EgoDexter~\cite{mueller2017real}, or HANDS17~\cite{yuan20172017}. Recently, partly motivated by AR/VR use cases where depth sensors are often unavailable due to high power consumption, the research community has largely switched to RGB or monochrome images, working on datasets such as the Stereo dataset~\cite{zhang20163d}, InterHand2.6M~\cite{moon2020interhand2}, FreiHAND~\cite{zimmermann2019freihand}, UmeTrack~\cite{han2022umetrack}, AssemblyHands~\cite{ohkawa2023assemblyhands}, and datasets with pose annotations of both hands and objects reviewed below.

\customparagraph{Datasets with objects only.}
Research on 6DoF object pose estimation and tracking has followed a similar path, starting off with custom monochrome datasets~\cite{roberts1963machine,murase1995visual} and later largely switching to RGB-D datasets such as LM~\cite{hinterstoisser2013model}, YCB-V~\cite{xiang2017posecnn}, and T-LESS~\cite{hodan2017tless}.
which are included in the BOP benchmark~\cite{hodan2018bop,hodan2020bop,sundermeyer2022bop,hodan2023bop}. The BOP benchmark currently includes twelve datasets in a unified format, offering 3D object models and training and test RGB-D images annotated with 6DoF object poses. The 3D object models are created manually or using KinectFusion-like systems for 3D surface reconstruction~\cite{newcombe2011kinectfusion}. The training images are real or synthetic (photo-realistically rendered with BlenderProc~\cite{denninger2019blenderproc,denninger2020blenderproc}) and all test images are real. Besides these instance-level datasets, the community also works on category-level RGB-D datasets such as Wild6D~\cite{fu2022category}, HouseCat6D~\cite{jung2022housecat6d}, and PhoCal\cite{wang2022phocal}. Recent methods started to focus again on estimating object pose from RGB-only images, using datasets such as OnePose~\cite{sun2022onepose} and HANDAL~\cite{guo2023handal}.

\customparagraph{Datasets with hands and objects.}
Many existing datasets include images of hands and objects (\eg, \cite{Fathi2011learning,Bullock2015yale,Bambach2015lending,shan2020understanding,zhang2022fine,Damen2022RESCALING}),
but only provide annotations in the form of 2D bounding boxes, segmentation masks, or action labels. Some datasets for 3D hand pose estimation (\eg, \cite{mueller2017real,zimmermann2019freihand,ohkawa2023assemblyhands}) include images of hands interacting with objects, but do not provide 6DoF object pose annotations.

The first dataset with ground-truth poses of both hands and objects was created by Sridhar~\etal \cite{RealtimeHO_ECCV2016} and offers 3014 exocentric RGB-D images of a hand manipulating a cube, manually annotated with fingertip positions and 6DoF poses of the cube. To avoid the manual annotation, which is tedious and not scalable, the FHPA dataset~\cite{Garcia2018first} used magnetic sensors attached to one hand and objects, noticeably affecting their appearance. This dataset includes 105K egocentric RGB-D images with ground-truth poses of a single hand and 4 objects.
The ObMan dataset~\cite{hasson2019learning} resorted to synthesizing images of hands grasping objects, with the grasps generated by an algorithm from robotics.

HO-3D~\cite{Hampali2020HO3D} was the first dataset with real images annotated by an optimization procedure that leverages multi-view RGB-D image streams and is almost fully automatic. The dataset offers 78K images from several exocentric cameras, showing 10 subjects and 10 objects. A similar annotation procedure was used for several subsequent datasets~\cite{huang2020h2o,hampali2022keypoint,chao2021dexycb,liu2022hoi4d, bhatnagar22behave}.
H2O~\cite{huang2020h2o} includes 572K egocentric multi-view RGB-D images of 4 subjects manipulating 8 objects.
H2O-3D\cite{hampali2022keypoint} provides 75K exocentric RGB images of 5 subjects manipulating 10 YCB objects~\cite{calli2015ycb}. 
DexYCB~\cite{chao2021dexycb} consists of 1000 clips of 3 seconds with the total of 582K RGB-D images,
recorded from 8 exocentric views and showing 10 subjects picking up 20 YCB objects with near-static grasps. HOI4D~\cite{liu2022hoi4d} includes 2.4M egocentric RGB-D images from over 4000 video sequences showing 9 subjects interacting with 800 different objects from 16 categories in 610 different indoor environments.
Besides rigid objects, this dataset contains articulated objects, but focuses on simpler scenarios with a single hand and a single object, and only includes single-view video sequences.
An RGB-D optimization procedure was also used in ContactPose~\cite{brahmbhatt2020contactpose} along with the information from thermal cameras for accurately annotating hand poses, while the object poses were annotated using optical markers. ContactPose includes 2.9M RGB-D images of 50 subjects grasping 25 household objects, however, the grasps are static, background green and all objects are blue (3D printed), which makes the images less realistic.

Similar to HOT3D, ground-truth poses of hands and objects in the recent ARCTIC dataset~\cite{fan2023arctic} were collected with a marker-based motion-capture system. This dataset includes 2.1M RGB images showing 10 subjects interacting with 11 articulated objects. The images were captured at 233K timestamps from 9 views, only one of which is egocentric (recorded with a mock-up of an egocentric device -- a camera mounted on a helmet).

\section{HOT3D dataset} \label{sec:dataset}

\noindent\textbf{833 minutes of recordings.}
The HOT3D dataset includes~egocentric, multi-view, synchronized data streams recorded with Project Aria~\cite{engel2023project} and Quest 3~\cite{Quest3}. Image streams contain 1.5M multi-view frames consisting of 3.7M images. Each frame from Aria consists of one RGB 1408$\times$1408 image and two monochrome 640$\times$480 images. Each frame from Quest~3 consists of two monochrome 1280$\times$1024 images. Intrinsic camera parameters and camera-to-world transformations are available for all images. All streams were recorded at 30 frames per second. Every recording from Aria also includes a 3D point cloud of the scene (from SLAM) and per-frame eye gaze information.
See App.~\ref{sec:appendix-aria} and \ref{sec:appendix-quest} for more details about Aria and Quest 3.

\customparagraph{3D mesh models of 33 objects.} The models were obtained by an in-house 3D object scanner and provide high-resolution geometry and PBR materials~\cite{mcdermott2018pbr}, which consist of metallic, roughness, and normal maps, and enable rendering of photo-realistic training images~\cite{hodan2019photorealistic,hodan2020bop}. The object collection includes household and office objects of diverse appearance, size, and affordances (Fig.~\ref{fig:models}).

\customparagraph{19 diverse subjects.} To ensure diversity, we recruited 19~participants with different hand shapes and nationalities. Hands of each participant were scanned by a custom 3D hand scanner and are provided in the UmeTrack~\cite{han2022umetrack} and MANO~\cite{mano:romero:tog17} formats.

\begin{figure}[t!]
  \centering
   \includegraphics[width=0.94\linewidth]{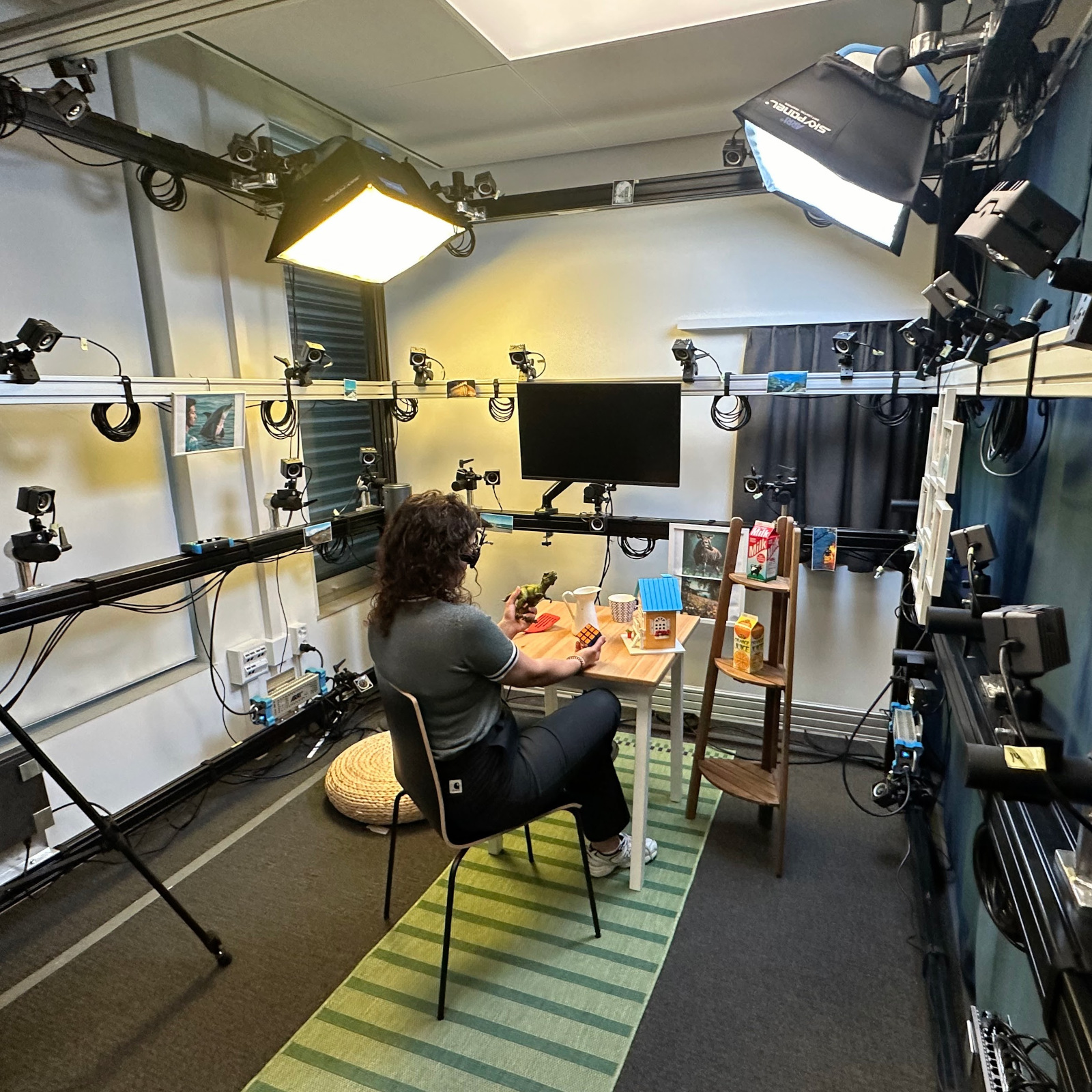}
   \caption{
   \textbf{Motion-capture lab.} 
   The HOT3D dataset was collected using a motion-capture rig equipped with a few dozens of infrared exocentric OptiTrack cameras and light diffuser panels for illumination variability.
   }
   \label{fig:rig}
\end{figure}

\customparagraph{4 everyday scenarios.}
Besides a simple inspection scenario, where subjects pick up, observe, and put down the objects, subjects were asked to perform typical actions in a kitchen, office, and living room. All scenarios were captured in the same lab equipped with scenario-specific furniture. In each recording, subjects were asked to interact with  up to 6 objects. To enhance diversity within the dataset, we regularly randomized various aspects such as lighting conditions, furniture placement, and decorative elements. The end result is a dataset comprising of 425 recordings, with 199 from Aria and 226 from Quest 3. Each recording has around 2 minutes.

\begin{figure}[t]
  \centering
  \includegraphics[width=1.0\linewidth]{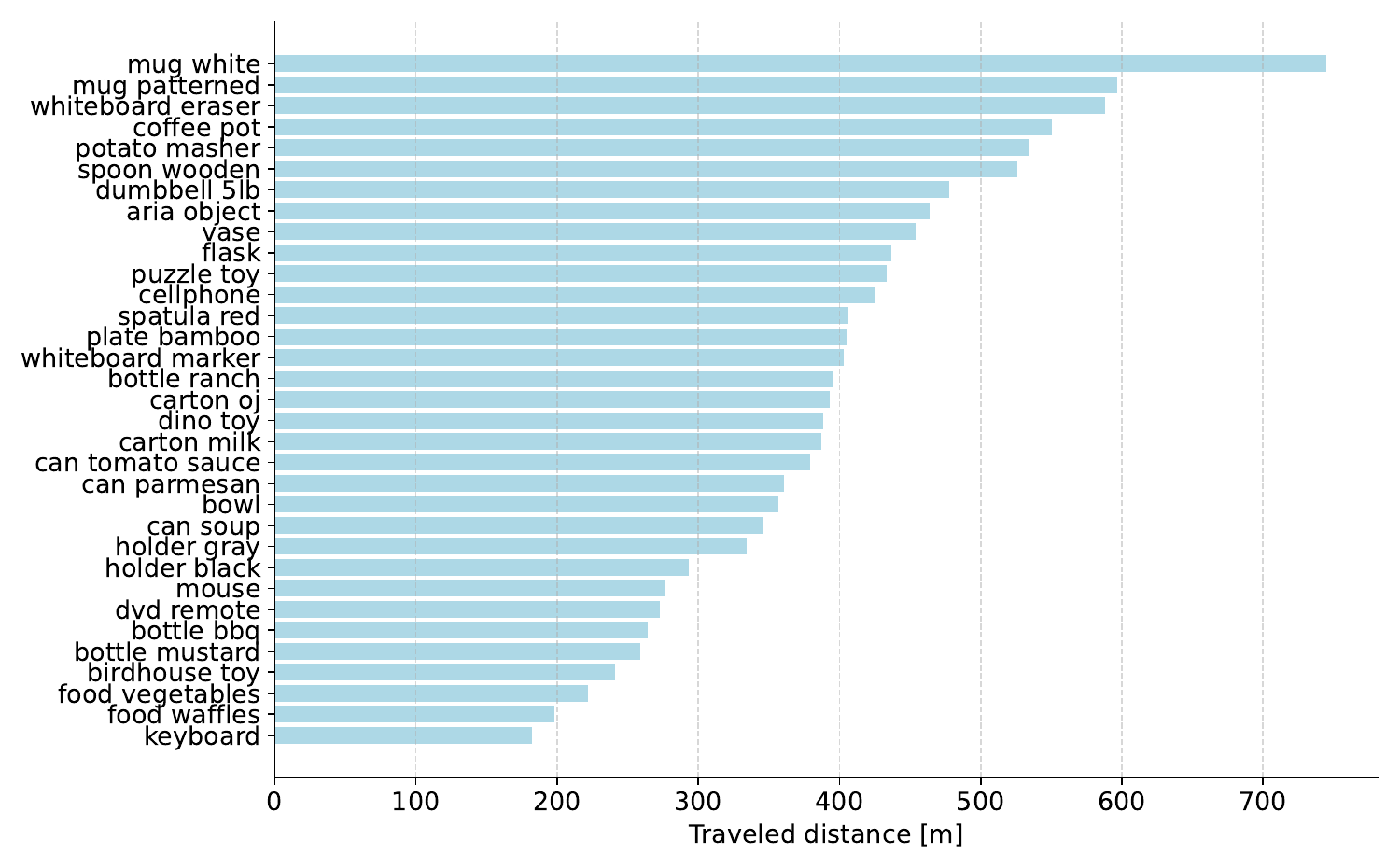} \vspace{-3ex}
   \caption{\textbf{Distances traveled by HOT3D objects.} In total, subjects moved the 33 objects over 13\,km. While objects like the keyboard and waffles were mostly resting, the white mug is a true explorer.
   }
   \label{fig:traveled_distances}
\end{figure}

\customparagraph{Ground-truth annotations.} Recordings are annotated with per-frame ground-truth poses of hands and objects obtained in a motion-capture lab shown in Fig.~\ref{fig:rig} and described in App.~\ref{sec:appendix-mocap}. Object and wrist poses are represented as 3D rigid transformation from the 3D model space to the scene space, and hand poses are represented in the UmeTrack~\cite{han2022umetrack} and MANO~\cite{mano:romero:tog17} formats (UmeTrack is more accurate while MANO more standard). Annotations in some frames may be missing or be of a lower quality. We visually inspected all recordings and flagged all frames with lower-quality poses. Out of 1.5M frames included in the dataset, 1.16M frames are fully annotated (\ie, ground-truth poses of all hands and objects are available) and passed our inspection. We release all 1.5M frames, which may be useful for unsupervised training, but also provide a mask of the valid frames. See Fig.~\ref{fig:traveled_distances} and \ref{fig:appendix:statistics:angular} for statistics of the ground-truth object poses.

\customparagraph{Training and test splits.} The training split of HOT3D includes recordings of 13 subjects (1M multi-view frames), and the test split includes recordings of the remaining 6 subjects (0.5 multi-view frames). Ground-truth pose annotations are publicly released only for the training split. Ground-truth annotations for the test split are accessible only by dedicated evaluation servers.

\customparagraph{Curated clips.} To facilitate benchmarking of various tracking and pose pose estimation methods, we also release 4117 curated clips extracted from the full recordings. 2969 clips come from the training split and 1148 from the test split. Each clip has 150 frames (5 seconds) which are all annotated with ground-truth poses of all modeled objects and hands and which passed our visual inspection. These clips are used in the public challenges introduced in Sec.~\ref{sec:challenges}.

\customparagraph{Object-onboarding sequences.} 
To enable benchmarking~model-free object tracking methods~\cite{sun2022onepose}, which learn new objects from reference images, and 3D object reconstruction methods~\cite{mildenhall2021nerf}, HOT3D also includes two types of onboarding sequences which show all possible views at each object: (1) sequences showing a static object on a desk, when the object is standing upright and upside-down, and (2) sequences showing an object manipulated by hands.
The static onboarding setup is suitable for NeRF-like reconstruction methods~\cite{mildenhall2021nerf}, while the latter is more practical for AR/VR applications yet more challenging~\cite{hampali2023hand}.
The ground-truth object poses are provided for all frames of static sequences but only for the first frames of dynamic sequences. This is to simulate real-world settings, where the poses can be easily obtained by SfM~\cite{schonberger2016structure} in the static setup, but are challenging to obtain in the dynamic setup. The ground-truth pose for the first frame of dynamic sequences is provided to define the canonical object space, which is necessary for evaluating 6DoF object tracking. Ground-truth hand poses are not provided for these sequences.

\section{Call for participation in public challenges} \label{sec:challenges}

The HOT3D dataset is used in two ongoing public challenges organized together with ECCV 2024 workshops: BOP Challenge 2024\footnote{\href{https://bop.felk.cvut.cz/challenges/bop-challenge-2024}{https://bop.felk.cvut.cz/challenges/bop-challenge-2024/}}, focused on model-free and model-based 2D/6D object detection, and Hand Tracking Challenge 2024\footnote{\href{https://github.com/facebookresearch/hand_tracking_toolkit?tab=readme-ov-file\#evaluation}{https://github.com/facebookresearch/hand\_tracking\_toolkit}}, focused on hand pose and shape estimation. To enable benchmarking methods for joint hand and object tracking, the two challenges use the same training and test HOT3D clips described in Sec.~\ref{sec:dataset}. We invite authors of relevant methods to participate in the challenges.

\clearpage

{\small
\bibliographystyle{ieee_fullname}
\bibliography{references}
}

\clearpage

\appendix
\section{Aria glasses} \label{sec:appendix-aria}

Project Aria~\cite{engel2023project} is an egocentric recording device in glasses form-factor created by Meta. It is designed as a \textit{research tool} for egocentric machine perception and contextualized AI research, and available to researchers across the world via \href{http://projectaria.com}{projectaria.com}.

\subsection{Device and sensors}
\label{appendix:aria:device}
Project Aria is built to emulate future AR- or smart-glasses catering to machine perception and egocentric AI rather than human consumption. It is designed to be wearable for long periods of time without obstructing or impeding the wearer, allowing for natural motion even when performing highly dynamic activities -- such as playing soccer or dancing. It has a total weight of 75g (compared to over 150g for a single GoPro camera), and fits just like a pair of glasses. 

Further, the device integrates a rich sensor suite that is tightly calibrated and time-synchronized, capturing a broad range of modalities. For HOT3D, \textit{recording profile 15} is used, which uses the following sensor configuration:
\begin{itemize}
    \item \textbf{One rolling-shutter RGB camera} recording at 30\,fps and $1408 \times 1408$ resolution. It is fitted with an F-Theta fisheye lens that covers a field of view of $110^{\circ}$. 
    \item \textbf{Two global-shutter monochrome cameras} recording at 30\,fps and $640 \times 480$ resolution. They provide additional peripheral vision, and are fitted with F-Theta fisheye lenses that cover a field of view of $150^{\circ}$.
    \item \textbf{Two monochrome eye-tracking cameras} recording at 10\,fps and $320 \times 240$ resolution.
    \item \textbf{Two IMUs} (800\,Hz and 1000\,Hz respectively), \textbf{a barometer} (50\,fps) and \textbf{a magnetometer} (10\,fps).
    \item \textbf{GNSS and WiFi} scanning was disabled for HOT3D. %
    \item \textbf{Audio} recording was disabled for HOT3D for privacy reasons. 
\end{itemize}
All sensor streams come with metadata such as precise timestamps and per-frame exposure times, and are made available in raw form as part of HOT3D. For convenience, we also include curated clips that suit the needs of the BOP and Hand Tracking challenges.

\begin{figure}[t]
    \vspace{-0.5ex}
    \centering
    \includegraphics[width=0.9\linewidth]{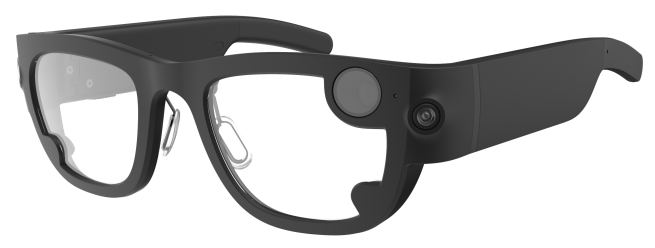}
    \caption{
        \textbf{Project Aria research glasses.}%
    }\vspace{-1.0ex}
    \label{fig:appendix:aria:device}
\end{figure}

\begin{figure}
    \centering
\begin{minipage}{0.50\linewidth}%
\includegraphics[width=\linewidth]{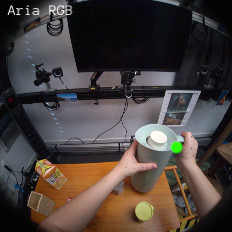}%
\end{minipage}
\begin{minipage}{0.235\linewidth}%
\includegraphics[width=\linewidth]{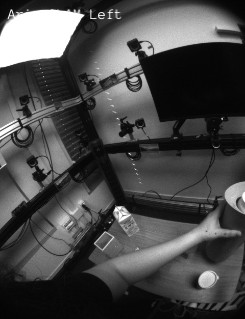}\\[0.5mm]%
\includegraphics[width=\linewidth]{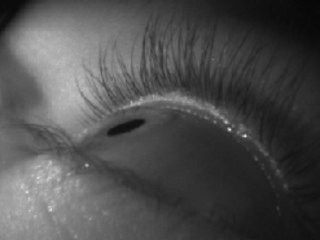}%
\end{minipage}
\begin{minipage}{0.235\linewidth}%
\includegraphics[width=\linewidth]{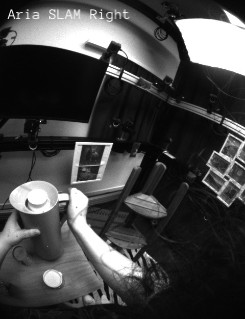}\\[0.5mm]%
\includegraphics[width=\linewidth]{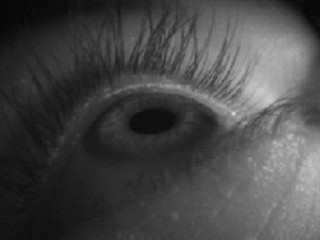}%
\end{minipage}\\%
\includegraphics[width=0.19\linewidth]{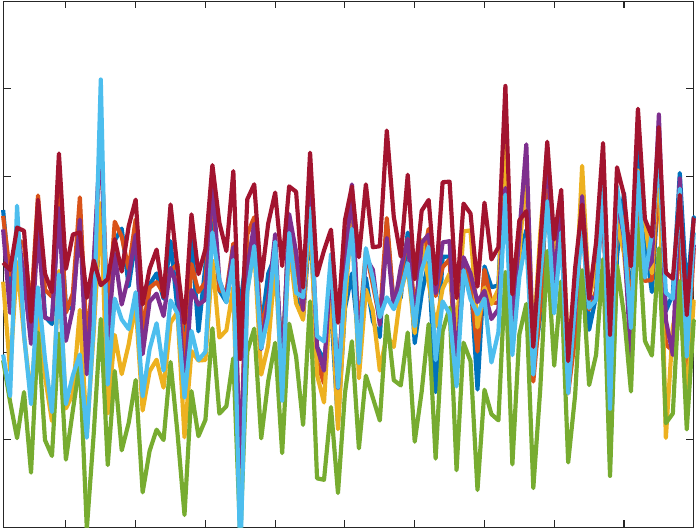}
\includegraphics[width=0.19\linewidth]{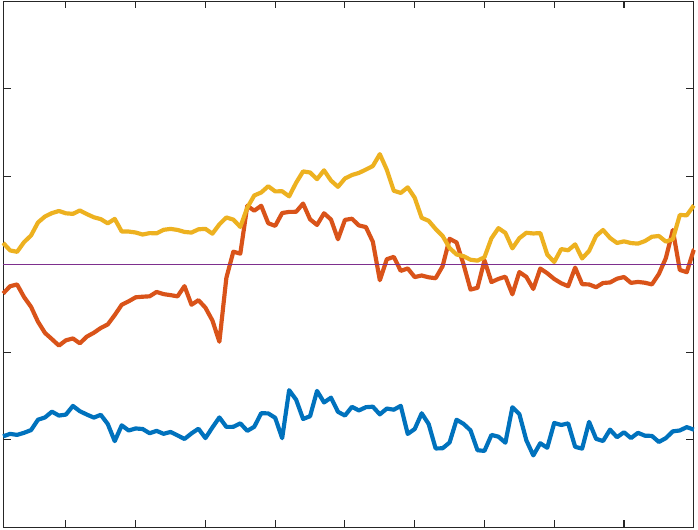}
\includegraphics[width=0.19\linewidth]{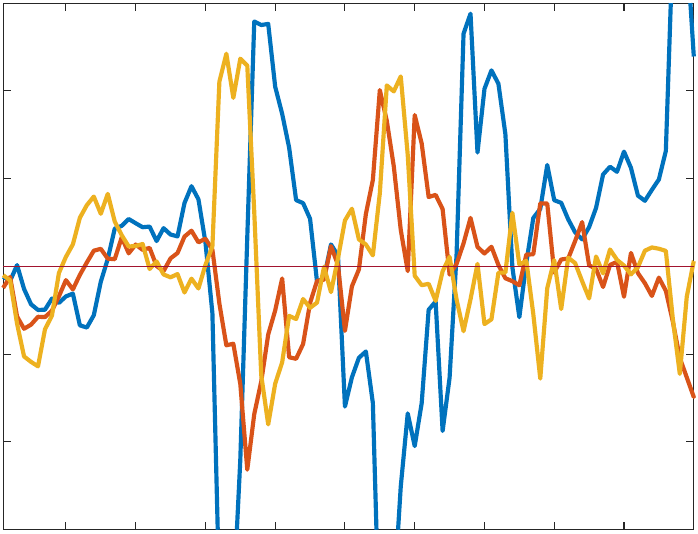}
\includegraphics[width=0.19\linewidth]{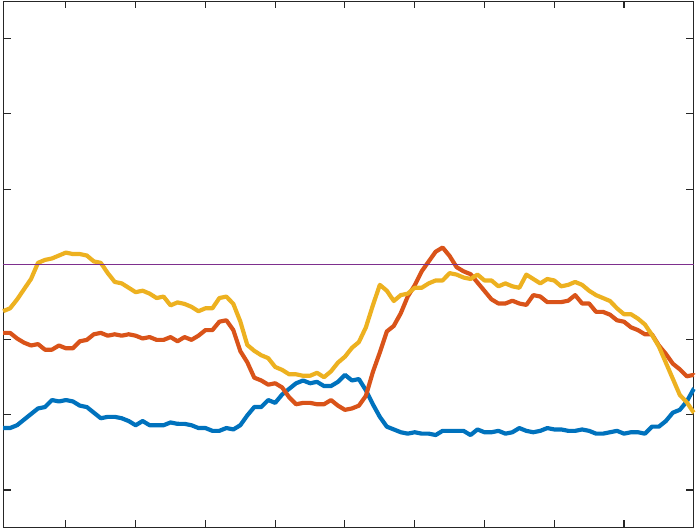}
\includegraphics[width=0.19\linewidth]{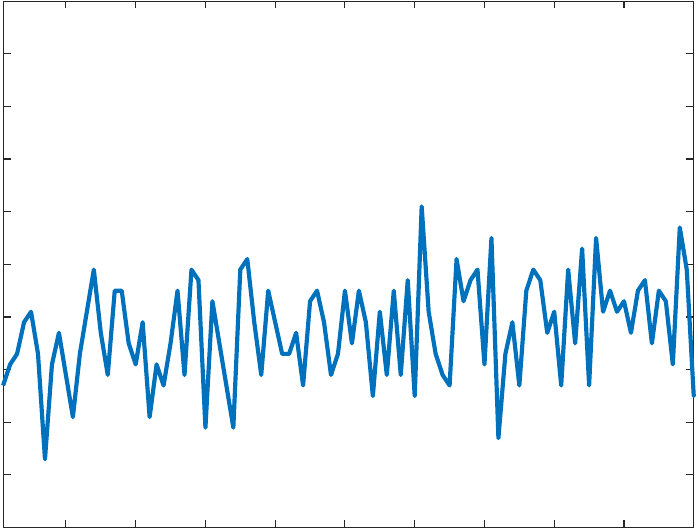}
    \caption{\textbf{Sensor streams recorded by the Project Aria device.} Top: RGB camera, left and right monochrome and eye cameras. Bottom: 10-second extracts from microphones, accelerometer, gyroscope, magnetometer and barometer respectively.}
    \label{fig:appendix:aria:sensors}
\end{figure}

\begin{figure}
    \centering
{\setlength{\fboxsep}{0pt}\setlength{\fboxrule}{0.5pt}
\fbox{\includegraphics[width=0.325\linewidth]{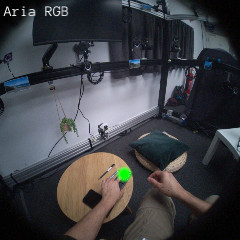}}\hfill%
\fbox{\includegraphics[width=0.325\linewidth]{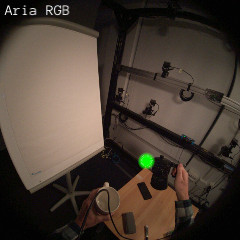}}\hfill%
\fbox{\includegraphics[width=0.325\linewidth]{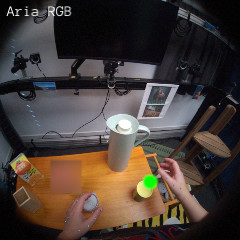}}\\[0mm]%
\fbox{\includegraphics[width=0.325\linewidth]{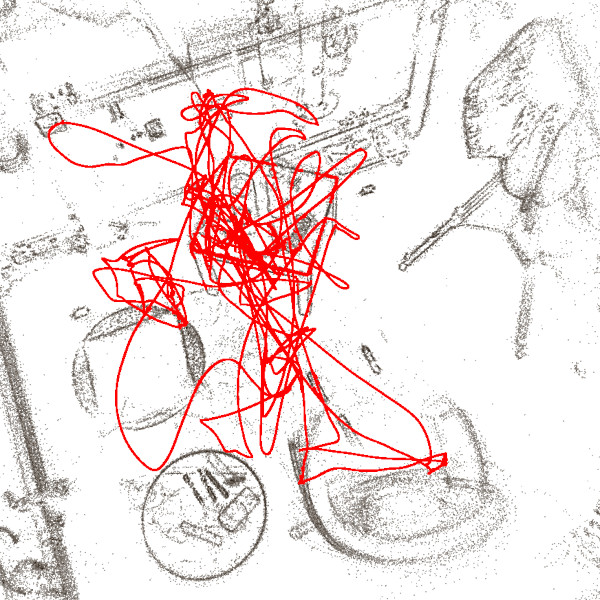}}\hfill%
\fbox{\includegraphics[width=0.325\linewidth]{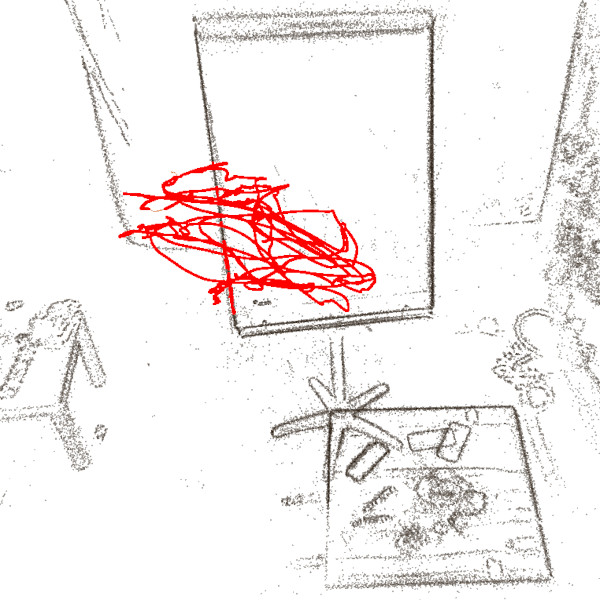}}\hfill%
\fbox{\includegraphics[width=0.325\linewidth]{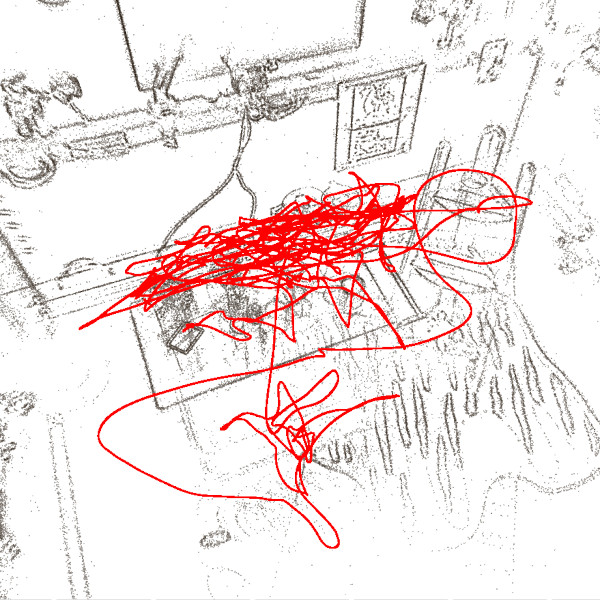}}\\[0mm]%
\fbox{\includegraphics[width=0.325\linewidth]{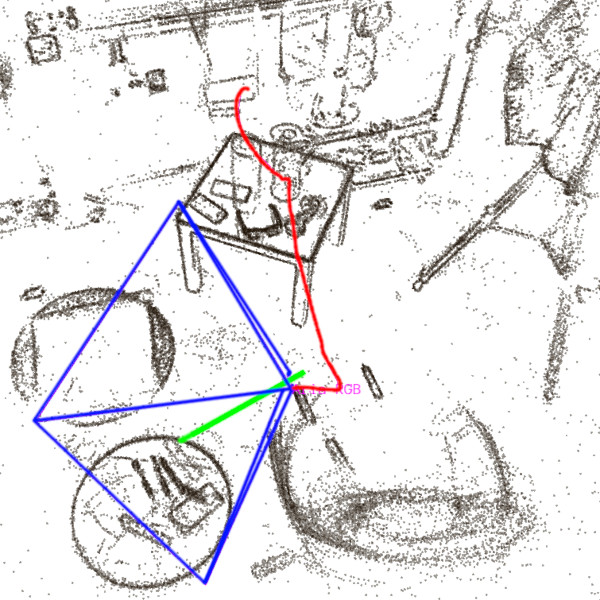}}\hfill%
\fbox{\includegraphics[width=0.325\linewidth]{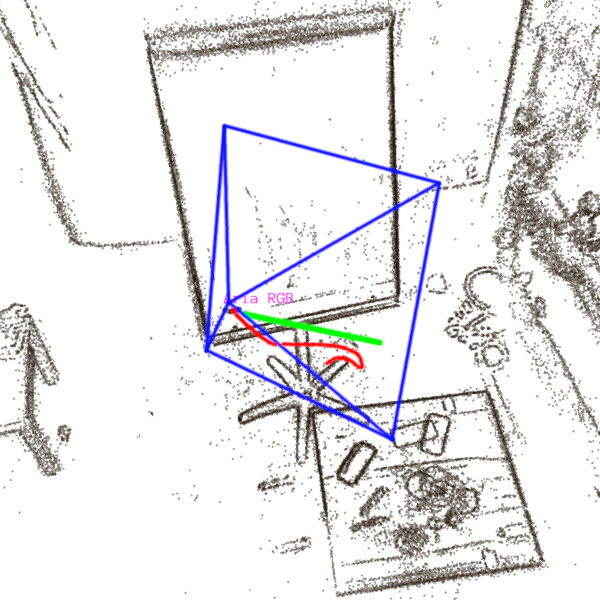}}\hfill%
\fbox{\includegraphics[width=0.325\linewidth]{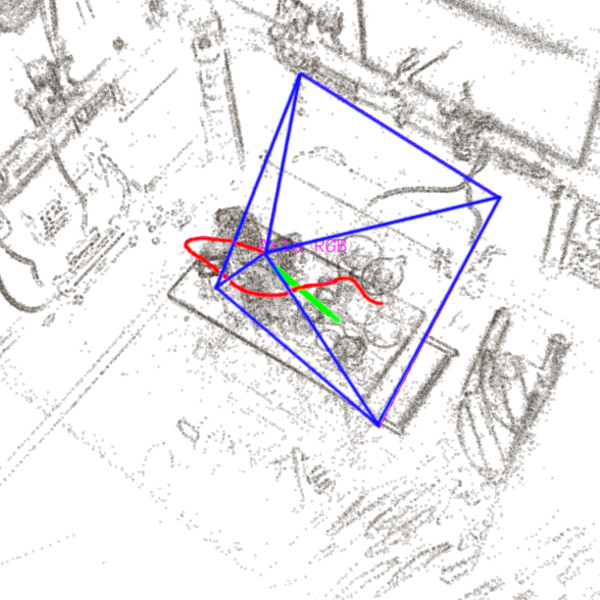}}%
}
    \caption{\textbf{Aria MPS output.} Shown is output for three recordings in a living room, office and kitchen scenario respectively (left to right). Top: RGB view and gaze (green dot). Middle: Point cloud and estimated egocentric camera trajectory for the full recording. Bottom: 3D view of a specific point in time, showing the RGB camera frustum (blue), gaze vector (green) and trajectory from the previous second (red).}
    \label{fig:appendix:aria:mps}
\end{figure}

\subsection{Machine Perception Services (MPS)}
\label{appendix:aria:mps}
Project Aria's machine perception service (MPS) provides software building blocks that simplify leveraging the different modalities recorded. These functionalities are likely to be available as real-time, on-device capabilities in future AR- or smart-glasses. We use the following core functionalities currently offered by Project Aria, and include their raw output as part of the dataset. See \cite{engel2023project} and the technical documentation\footnote{\href{https://facebookresearch.github.io/projectaria_tools/docs/intro}{https://facebookresearch.github.io/projectaria\_tools/docs/intro}} for more details. 

\customparagraph{Calibration.} All sensors are intrinsically and extrinsically~calibrated, and tiny deformations due to temperature changes or stress applied to the glasses frame are further corrected by time-varying online calibration from MPS.

\customparagraph{Aria 6\,DoF localization.} Every recording is localized precisely and robustly in a common, metric, gravity-aligned coordinate frame, using a state-of-the-art VIO and SLAM algorithm. This provides millimeter-accurate 6\,DoF poses for every captured frame and high-frequency (1\,kHz) motion in-between frames. 

\customparagraph{Eye gaze.} The gaze direction of the user is estimated as two outward-facing rays anchored approximately at the wearer's eyes, allowing to approximately estimate not only the direction the user is looking in, but also the depth their eyes are focused on. We use an optional eye gaze calibration procedure, where the mobile companion app directs the wearer to gaze at a pattern on the phone screen while performing specific head movements. This information was then used to generate a more accurate eye gaze direction, personalized to the particular wearer.

\customparagraph{Point clouds.} A 3D point cloud of static scene elements is triangulated from the moving Aria device, using photometric stereo over consecutive frames or left/right SLAM camera. Points are added causally over time, and will include points on any object that is observed while static for several seconds. The output contains both the 3D point clouds as well as the raw 2D observations of every point in the camera images it was triangulated from.

\subsection{Processing summary} 
All Aria recordings are anonymized in a very first step, using the public EgoBlur \cite{raina2023egoblur} model and following Project Aria's responsible innovation principles. 

Then, the MPS pipeline is invoked for each full Aria recording, which are typically about 2 minutes long and include many instances of hand-object interactions with different objects.
Next, we 7DoF-align the MPS output with the OptiTrack coordinate frame (App.~\ref{sec:appendix-mocap}). In total, we have processed 199 Aria recordings with a total length of 391 minutes. See Sec.~\ref{sec:dataset} for additional statistics.

\subsection{Tools and ecosystem}
\label{appendix:aria:tools}
Technical documentation and open-source tooling for Aria recordings and MPS output is available on GitHub\footnote{\href{https://github.com/facebookresearch/projectaria_tools}{https://github.com/facebookresearch/projectaria\_tools}} and the associated documentation page\footnote{\href{https://facebookresearch.github.io/projectaria_tools/docs/intro}{https://facebookresearch.github.io/projectaria\_tools/docs/intro}}. It includes both python and C++ tools to convert, load, and visualize data; as well as sample code for common machine perception and 3D computer vision tasks. 

\section{Quest 3 headset} \label{sec:appendix-quest}

\begin{figure}[t]
    \centering
    \includegraphics[width=0.9\linewidth]{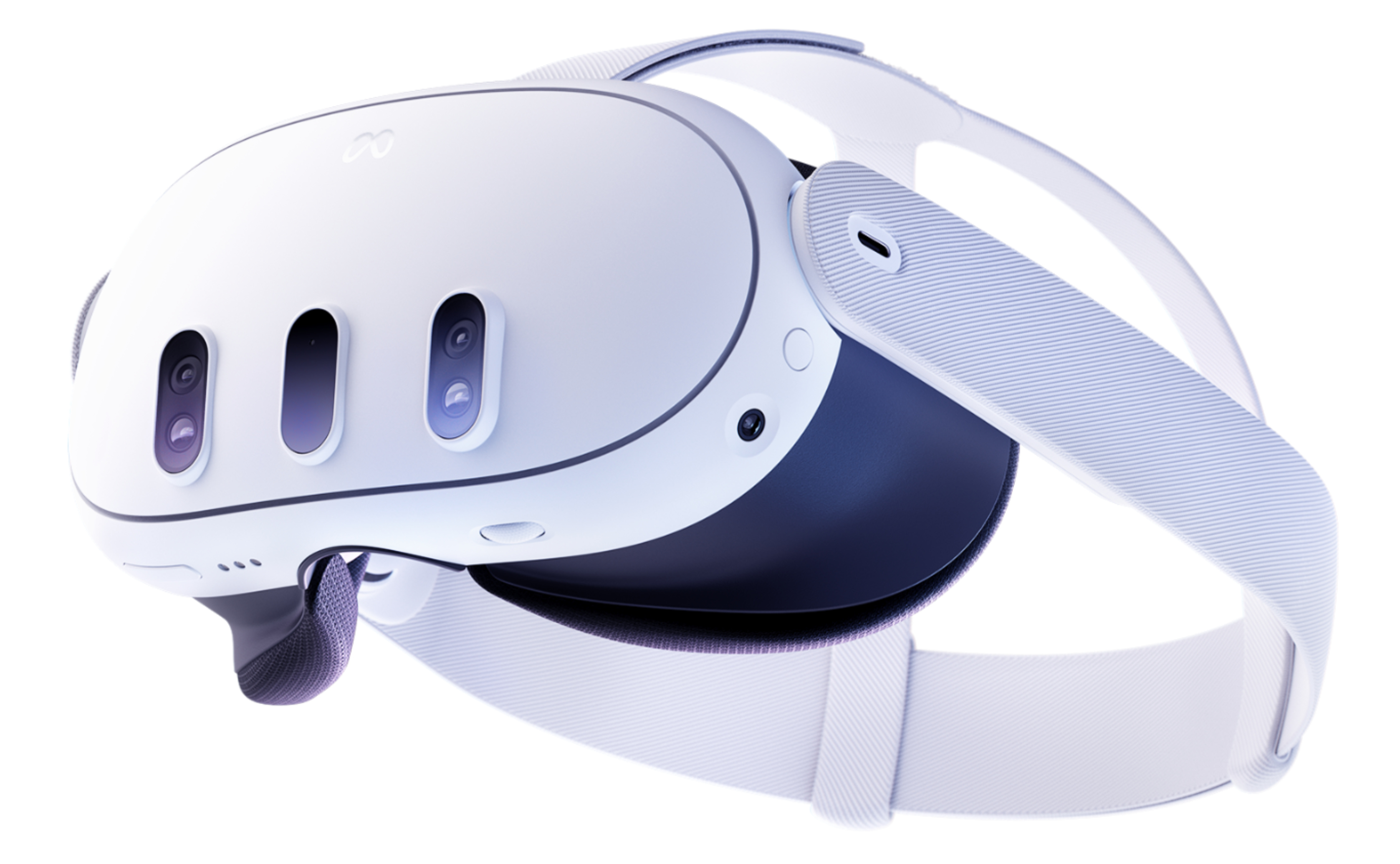}
    \caption{
        \textbf{Meta Quest 3 headset for virtual and mixed reality.}
    }\vspace{1.0ex}
    \label{fig:appendix:quest3:device}
\end{figure}

\begin{figure}[t]
    \centering
    \setlength{\tabcolsep}{1pt} %
    \renewcommand{\arraystretch}{0.6} %

    \begin{tabular}{cc}
        \includegraphics[width=0.492\linewidth]{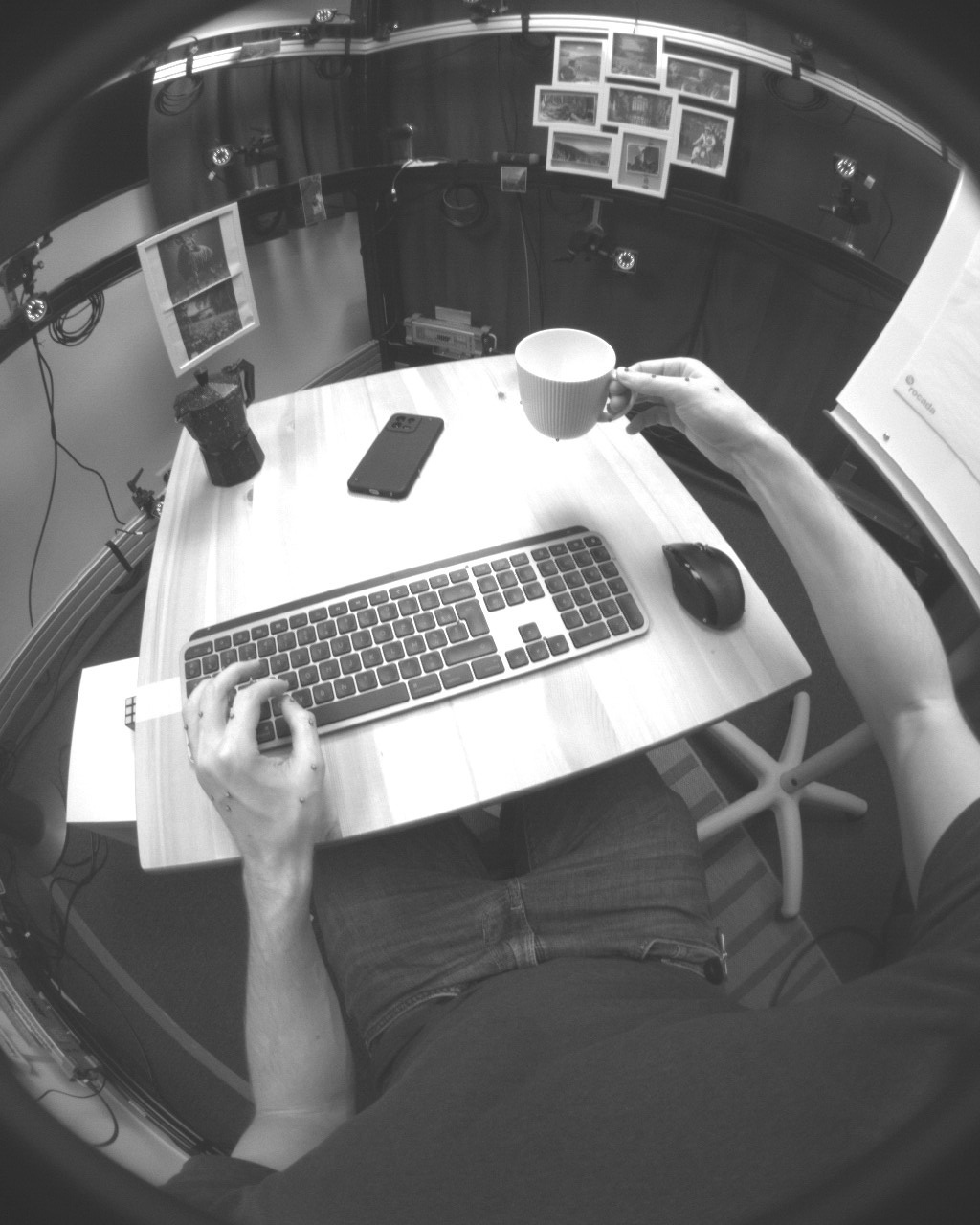} &
        \includegraphics[width=0.492\linewidth]{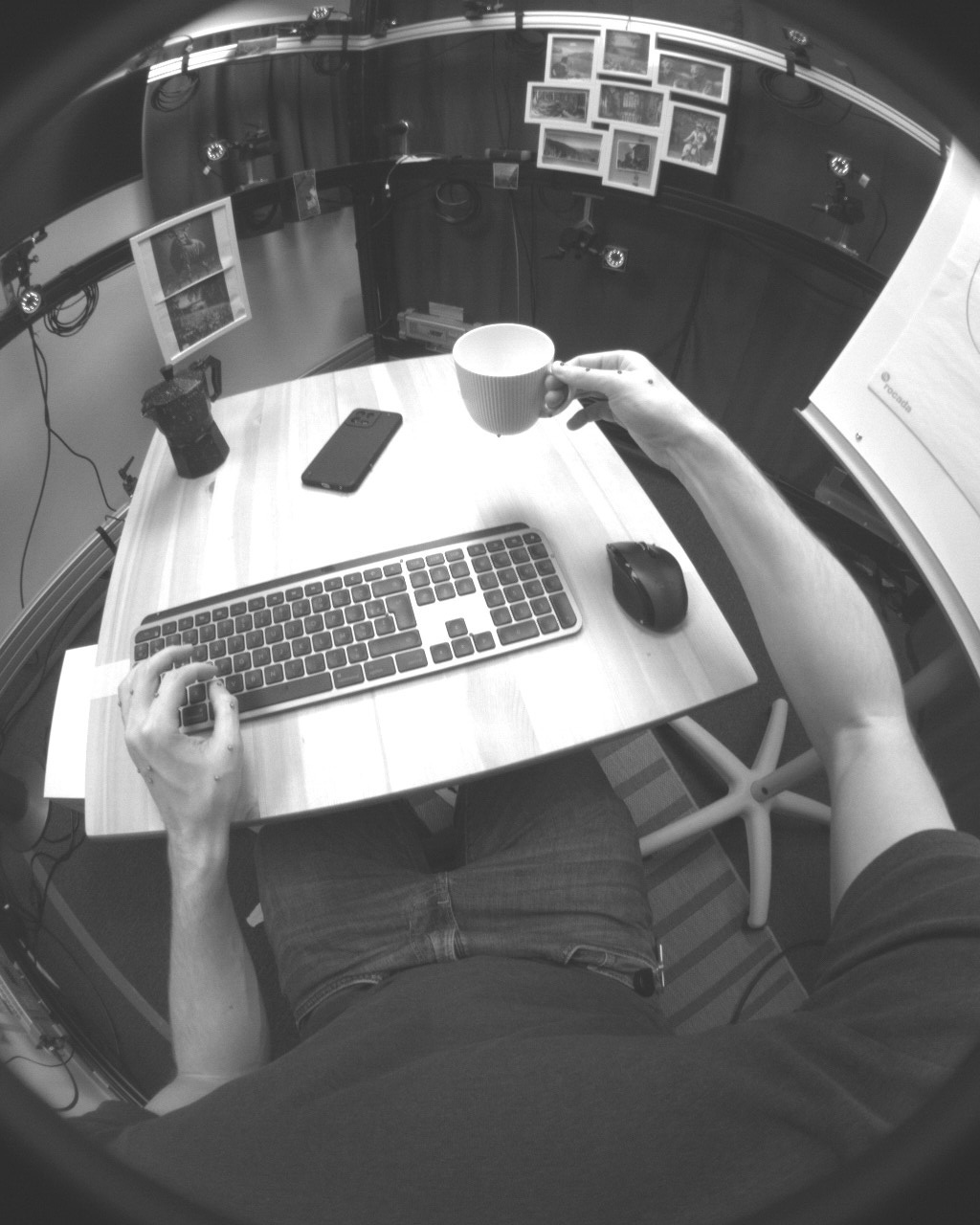} \\
        \includegraphics[width=0.492\linewidth]{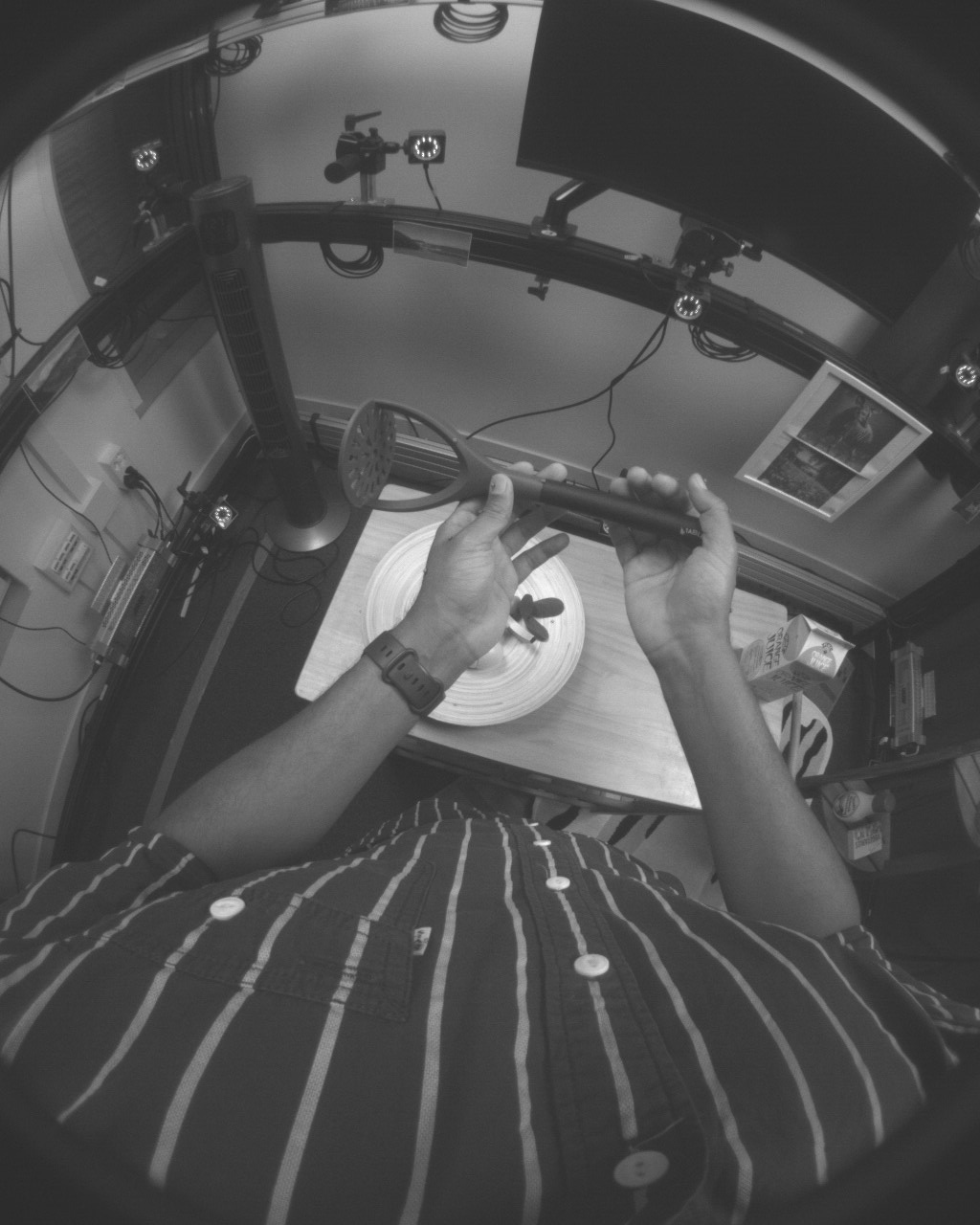} &
        \includegraphics[width=0.492\linewidth]{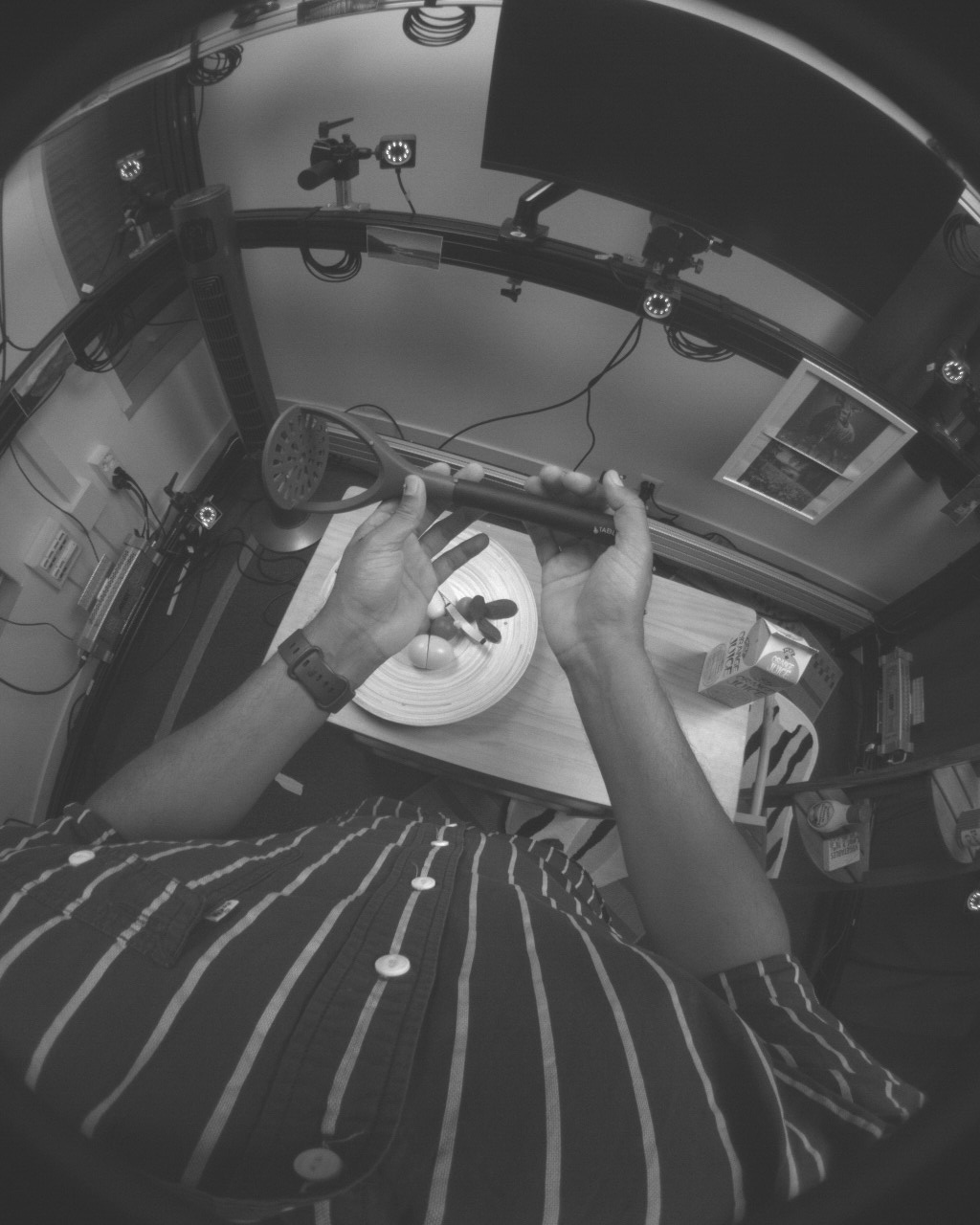}
    \end{tabular}
    \vspace{-0.7ex}
    \caption{
        \textbf{Sample images from Quest 3.} Shown are synchronized images from the two front Quest 3 cameras used for the HOT3D collection.
    }
    \label{fig:appendix:quest3:images}
\end{figure}

Quest 3~\cite{Quest3}, shown in Fig.~\ref{fig:appendix:quest3:device}, is the latest production headset from Meta for virtual- and mixed-reality experiences.
For the HOT3D data collection we used an internal developer version of the Quest 3 headset. This version has four global-shutter monochrome cameras with fisheye lenses, 1280x1024\,px image resolution, 18\,PPD (Pixels Per Degree), and recording at 30\,fps.
Two of the cameras are on the front side of the headset, roughly aligned with eyes, and two on the sides. HOT3D only includes images from the two front cameras as those capture the relevant scene part (the two side cameras are useful for applications like SLAM). Example images are in Fig.~\ref{fig:appendix:quest3:images}.
Data from other sensors present in the consumer version of Quest~3, including a gyroscope and an accelerometer, were not recorded. The intrinsic and extrinsic parameters of the headset cameras were calibrated with a ChArUco board. Both the headset and the board were attached a set of optical markers and tracked by the motion-capture system described in App.~\ref{sec:appendix-mocap}, which allowed to estimate camera-to-headset transformations. At recording time, the headset pose was still tracked by the motion-capture system and used to calculate per-frame camera-to-world transformations.

\section{Marker-based motion capture} \label{sec:appendix-mocap}

The poses of hands and objects were tracked using optical markers attached on their surface. For both hands and objects we used 3\,mm markers with an adhesive layer at their bottom. Such markers are small enough not to influence hand-object interactions. Each hand was attached 19 markers and each object around 10. The marker locations were then semi-automatically registered to 3D models of hands and objects obtained by custom 3D scanners.

At recording time, the optical markers were tracked by multiple infrared OptiTrack
cameras attached on a rig shown in Fig~\ref{fig:rig}. The intrinsic and extrinsic parameters of the infrared cameras were calibrated before every capturing session.
Hand poses were calculated by fitting the participant's UmeTrack hand model~\cite{han2022umetrack} to the tracked optical markers, as in~\cite{MocapHT_Siggraph2018}. Object poses were estimated by aligning the tracked markers to their registered 3D locations in the model coordinate frame. To achieve reliable tracking, it was important to ensure that the marker constellation on each object is sufficiently distinct. Data frames from different sources were synchronized with SMPTE timecode.

\section{Object orientation statistics} \label{sec:appendix-statistics}

\begin{figure}[h!]
    \centering
\begin{minipage}{\linewidth}%
\includegraphics[width=\linewidth]{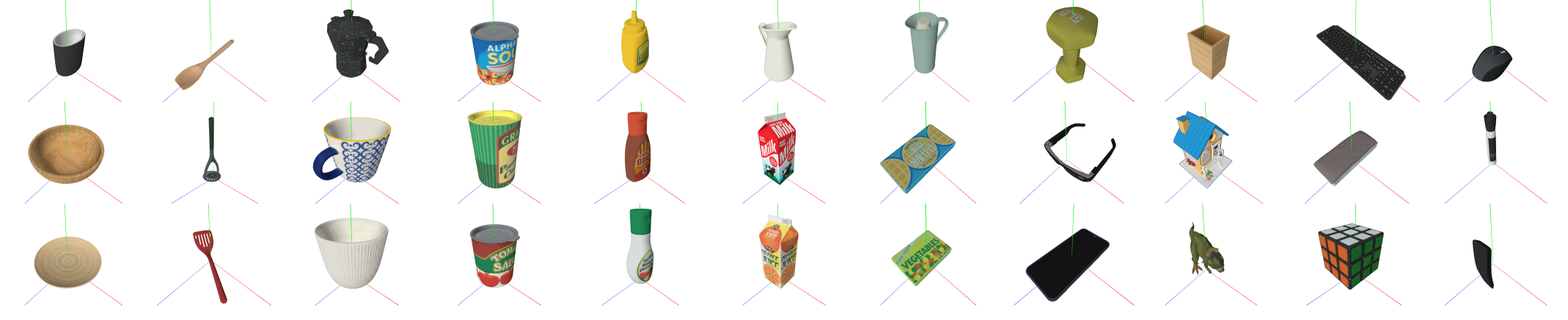}%
\end{minipage}
\begin{minipage}{\linewidth}%
\includegraphics[width=\linewidth]{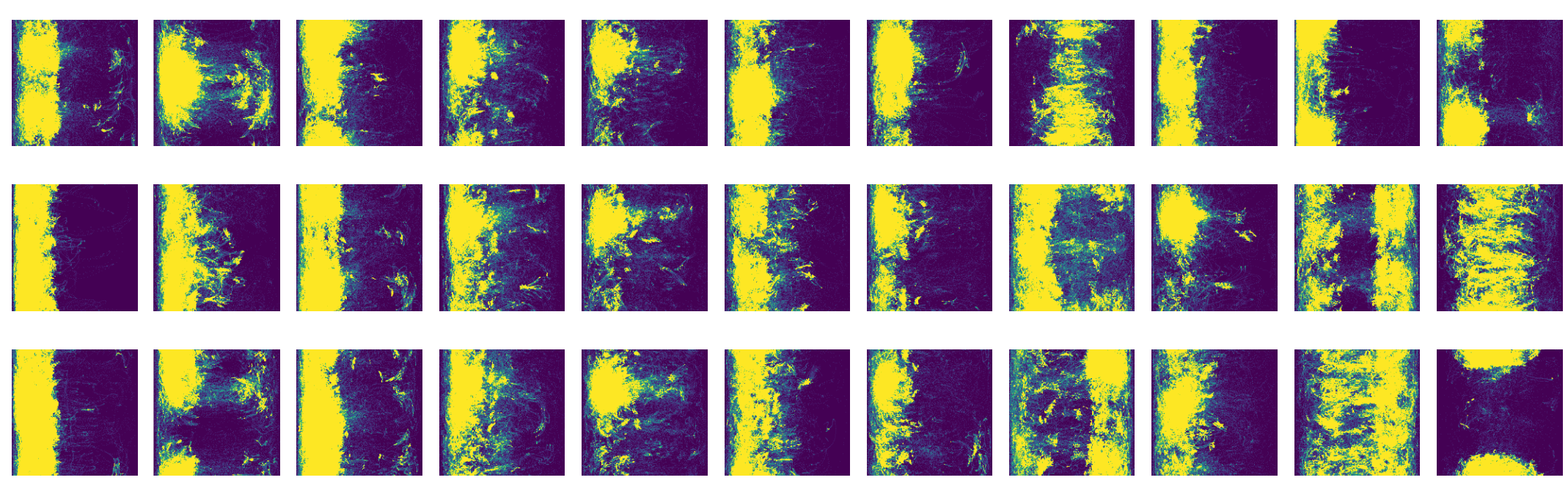}%
\end{minipage}
\caption{\textbf{Object orientation statistics.} Top: 3D object models in their canonical poses. Bottom: Distribution of azimuth and elevation angles under which the objects are observed across the dataset. The vertical axis is the azimuth angle $[0^\circ,\;360^\circ]$ (angle along the green axis), and the horizontal axis is the elevation angle $[-90^\circ,\;90^\circ]$ (angle \wrt the plane defined by the red and blue axes).}
\label{fig:appendix:statistics:angular}
\end{figure}

When recording HOT3D, we asked subjects to naturally interact with the objects.
Consequently, orientation distributions of the observed objects  (Fig.~\ref{fig:appendix:statistics:angular}) reveal clear object-specific pose biases, which may be useful as prior information at inference time (we see that the bowl tends to be seen upright, the birdhouse from the front and upright, \etc).

\end{document}